\pdfoutput=1
\documentclass[aps,pre,twocolumn,a4paper,superscriptaddress,floatfix]{revtex4-2}
\usepackage{amsmath,amssymb,bm}
\usepackage{graphicx}
\graphicspath{{figures/}}
\usepackage{xcolor}
\usepackage[colorlinks=true,linkcolor=blue!50!black,citecolor=blue!50!black,urlcolor=blue!50!black]{hyperref}

\newcommand{\kHA}{k_{\mathrm{HA}}}
\newcommand{\kAH}{k_{\mathrm{AH}}}

\begin{document}

\title{Skill erosion, individual memory and the reversibility of AI adoption}
\author{Renaud Lambiotte}
\affiliation{Mathematical Institute, University of Oxford, Oxford, United Kingdom}
\affiliation{Institute for New Economic Thinking, University of Oxford, Oxford, United Kingdom}
\affiliation{Complexity Science Hub, Vienna, Austria}
\date{\today}

\begin{abstract}
Generative AI can improve immediate performance but may erode the human skills needed for future work. We consider a population of agents who perform a task either with their own skill or with the help of AI. Agents tend to imitate more successful peers, and their skills evolve in time:
improving with practice and eroding with reliance on AI. We study, analytically and with simulations, the collective consequences of AI adoption. We show that the mechanism of skill erosion makes adoption bistable and that 
 the long-term behaviour of the population depends on how skills are distributed: populations with the same adoption and mean skill can reach opposite asymptotic states, depending on how expertise is shared between AI users and non-users. A deterministic description that follows the mean skills of the two groups and the variance of skills among non-users predicts the outcome of individual runs as accurately as the simulations allow, whereas the means alone are not sufficient. We also consider the situation when the benefit of AI improves in time. In that case, adoption can jump abruptly, and a temporary increase in benefit, such as in a period of low prices, can leave a population relying on AI after the benefit has returned to its original value. Skill erosion therefore creates a history-dependent switching cost that can make widespread AI adoption difficult to reverse.
\end{abstract}
\maketitle

\section{Introduction}\label{sec:intro}

Generative artificial intelligence (AI) is rapidly becoming part of how people write, code, learn and make decisions. It can improve immediate performance in a wide range of tasks, but it may also reduce performance in two ways. At the individual level, relying on AI may reduce the practice that maintains the ability of people to work independently, as AI assistance can help with a task yet reduce subsequent independent performance~\cite{liu}. Such effects are not specific to AI. Offloading a cognitive task to a tool~\cite{risko2016} can let the corresponding skill decline~\cite{bainbridge1983}. For instance, habitual use of satellite navigation is associated with a decline of spatial memory, including among expert drivers~\cite{dahmani2020,ying2024}, and automating flight planning degrades the cognitive skills that it replaces~\cite{volz2020}. Similar deskilling has been reported among experienced endoscopists after routine AI assistance~\cite{budzyn2025}. At the population level, the value of AI-assisted work may decrease as more people rely on the same tools~\cite{pedreschi2025human}: AI-assisted outputs become more similar, reducing the diversity of ideas and content~\cite{doshi,meincke}, much as navigation services that give the same advice to many drivers concentrate traffic on the same roads~\cite{cornacchia2026traffic}.

As with past innovations, the adoption of AI is fundamentally a social process~\cite{karsai2014complex}. A software developer adopting a code assistant, a student solving exercises with a chatbot or an analyst drafting reports with a language model all observe how well these tools serve their peers before adopting them. Models of social learning and evolutionary game theory therefore provide a natural framework for such dynamics~\cite{szabo,castellano}.
AI adoption differs from classical models in one important way: the payoff of a strategy does not depend on the current strategies of the agents alone, but also on the history of the individual who plays it. Indeed, each individual carries an internal state, their skill, which evolves with their own behavioural history: using AI changes the skill of the user, and therefore the payoff of their future decisions. Consider two AI users who behave identically today: one has just adopted AI and retains the expertise of years of independent practice, whereas the other has relied on AI for a long time and has lost it. We call this dependence of an agent's skill, and hence of the payoff of their next choice, on their own past actions \emph{individual memory}. 

Recent work has shown that the mechanism of skill erosion can make the adoption of AI bistable and history-dependent~\cite{frigerio,park2026enrichment,sole2026cognitive}. In a model closely related to ours \footnote{The present work was developed independently of Ref.~\cite{frigerio}, which was posted while this manuscript was being written. Some of our results, indicated in the text, were also obtained there in a weak-selection approximation, in particular the rate equations, the role of the skill of non-users in adoption and the stationary skill of non-users.}, Frigerio et al.~\cite{frigerio} showed that skill erosion and the loss of distinctiveness of AI-assisted work compete to produce two states, one of low adoption and high skill and one of high adoption and low skill. The region in which both states are stable narrows as behaviour becomes more persistent, and skills then polarise in ways hidden by population averages. Population-level models have also produced tipping points and irreversible dependence on AI, either by letting the mean capability of a population coevolve with its reliance on AI~\cite{park2026enrichment}, or by letting users move between uncoupled, coupled and dependent states, as in an epidemic~\cite{sole2026cognitive}. 
Here, we ask how individual memory shapes the collective dynamics of AI adoption: under what conditions, and from which states, does a population adopt AI, or does it return to independent work? Note that the second question is the richer one, as skill erosion leaves the threshold of adoption unchanged, but acts on the possibility of return. We show that the answer depends on individual memory, i.e. on the skill that each agent carries as a result of their own history.

To address this overarching question, we investigate a model with three mechanisms: skill erosion, random exploration and crowding, i.e.\ the loss of distinctiveness of AI-assisted work as more people use it. 
We then investigate it via three angles, each time considering the simplest version of the model that produces the effect under study. After introducing the model (Sec.~\ref{sec:model}) and the equations that describe the population (Sec.~\ref{sec:exact}), we first derive the thresholds of adoption and return (Sec.~\ref{sec:return}). When skill erosion is the only feedback and agents change behaviour only by imitation, these thresholds follow analytically from whether a lone agent with a different  behaviour of that of the population can spread. They are asymmetric: a population of experts adopts AI only if AI-assisted work earns more than expert work, whatever the persistence of behaviour, whereas a population relying on AI returns only if a lone non-user regains enough skill before switching back, a threshold set by how quickly skills are relearned rather than by how quickly they erode. 

Second, we ask what determines the fate of a population (Sec.~\ref{sec:fate}). Populations with the same adoption and the same mean skill can end in opposite states, depending on how skill is shared between those who still work without AI and those who rely on it, and we show that the outcome is predicted by following the mean skills of the two groups and the variance of skills among non-users. 

Third, we ask how long an advantage of AI must last to lock a population in (Sec.~\ref{sec:howlong}). 
To investigate this problem, we  incorporate the mechanism of random exploration, to ensure that the population does not end in an absorbing state at stationarity. 
Our results show that skill erosion acts as a switching cost, a classic source of lock-in in economics~\cite{arthurlockin,klemperer}, but one that is paid when leaving AI rather than when adopting it, through the user's own eroded skill, and that grows the longer AI has been used.

Finally, we investigate the effect of an additional feedback on the collective dynamics, crowding, and determine  how it modifies the conclusions (Sec.~\ref{sec:crowding}). We discuss our results in Sec.~\ref{sec:discussion}. 

\begin{figure}[t]
\centering
\includegraphics[width=\columnwidth]{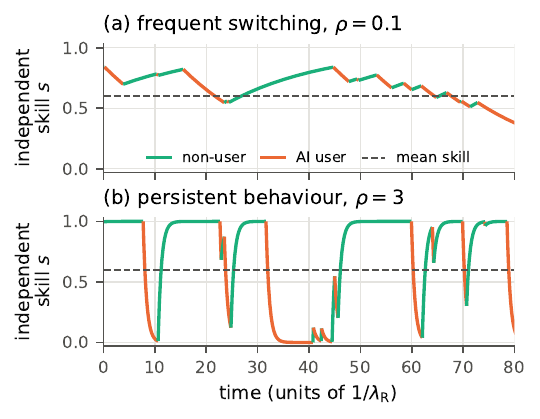}
\caption{\textbf{Individual memory.} Illustrative skill trajectories of a single agent, coloured by behaviour, when behaviour changes often compared with the time skills need to adapt (a, $\rho=0.1$) and when behaviour is persistent (b, $\rho=3$); $\rho$, Eq.~\eqref{eq:rho}, compares the rate at which skills adapt with the rate at which behaviour changes. For illustration, the switching rates are constant and chosen so that the agent uses AI $40\%$ of the time. The agent has the same mean skill in both cases (dashed line), but skill stays close to the mean in (a) and alternates between expertise and deep erosion in (b).}
\label{fig:model}
\end{figure}

\section{Model}\label{sec:model}

We consider a population of $N$ individuals, or agents. Each agent $i$ has a behaviour $\sigma_i\in\{\mathrm H,\mathrm A\}$, working either without AI (H, for human) or with AI (A), and an independent skill $s_i\in[0,1]$, the quality of the work they can produce without AI. We denote by $x$ the fraction of AI users, which we call adoption. The payoff of each behaviour is
\begin{equation}
\pi_{\mathrm H}(s)=s,\qquad \pi_{\mathrm A}=b-c\,x .
\label{eq:returns}
\end{equation}
Work without AI is worth the skill of the agent, whereas AI-assisted work is worth a benefit $b$, reduced by a crowding term $c\,x$ that captures the loss of distinctiveness as more people use the same tools~\cite{doshi,meincke}. For simplicity, we assume that the payoff of AI-assisted work does not depend on skill. We discuss the scenario of AI tools that complement expertise in Appendix~\ref{sec:app-abm}.

Between changes of behaviour, skills evolve as
\begin{equation}
\dot s_i=\begin{cases}\lambda_{\rm H}\,(1-s_i) & \text{if }\sigma_i=\mathrm H,\\ -\lambda_{\rm A}\,s_i & \text{if }\sigma_i=\mathrm A .\end{cases}
\label{eq:skill}
\end{equation}
Practice pulls the skill of an agent towards $1$ at rate $\lambda_{\rm H}$, while reliance on AI lets it erode towards $0$ at rate $\lambda_{\rm A}$. Unless stated otherwise, we take the same rate for learning and erosion, $\lambda_{\rm H}=\lambda_{\rm A}=\lambda$, and we discuss their distinct roles in Sec.~\ref{sec:rates}.

Agents revise their behaviour at rate $\lambda_{\rm R}$. At each revision, an agent explores with probability $\mu$, i.e.\ adopts one of the two behaviours chosen at random, whatever their current behaviour. Otherwise, the agent observes a randomly chosen peer $j$ and, if their behaviours differ, copies $j$ with probability
\begin{equation}
P(i\leftarrow j)=\tfrac12+\tfrac{\beta}{2}\big(\pi_j-\pi_i\big),
\label{eq:imitation}
\end{equation}
that is a linear pairwise comparison rule~\cite{traulsen2005coevolutionary,traulsen}. The parameter $\beta$ sets the strength of selection, i.e.\ how much the probability of copying depends on the difference of payoffs; for $\beta\to0$, copying is nearly random. Importantly, agents compare \emph{realised} payoffs, which depend on the individual memory of their peer, and do not anticipate the payoff they would obtain after switching: an AI user who copies a skilled non-user then earns their own, possibly eroded, skill, not that of the peer they copied. Equation~\eqref{eq:imitation} is the weak-selection expansion of the Fermi rule, but we use it at finite $\beta$, which only requires that it define a probability; this holds for all the parameters that we use (Appendix~\ref{sec:app-rate}).

The model has two timescales, the time $1/\lambda_{\rm R}$ between revisions of behaviour and the time $1/\lambda$ over which skills adapt. We compare them through the ratio
\begin{equation}
\rho=\frac{2\lambda}{\lambda_{\rm R}(1+\mu)}\simeq\frac{2\lambda}{\lambda_{\rm R}}.
\label{eq:rho}
\end{equation}
The denominator, $\lambda_{\rm R}(1+\mu)/2$, is the total rate at which an agent switches behaviour when selection is absent ($\beta=0$). Small $\rho$ means that agents typically switch behaviour before their skills have adapted, whereas large $\rho$ means that behaviour persists long enough for skills to approach their targets. When learning and erosion proceed at different rates, we write $\rho_{\rm H}$ and $\rho_{\rm A}$ for Eq.~\eqref{eq:rho} with $\lambda=\lambda_{\rm H}$ and $\lambda=\lambda_{\rm A}$.

Figure~\ref{fig:model} illustrates the two regimes for a single agent: when the agent changes behaviour often, their skill stays close to its mean, whereas when their behaviour is persistent, it alternates between expertise and deep erosion. The correlation between current behaviour and skill, through which individual memory shapes collective behaviour, is thus strongest for large $\rho$, when skill closely follows behaviour. As $\rho\to0$, an agent switches many times before their skill can adapt, so that their current behaviour carries little information about their skill.

Linear payoffs, linear relaxation of skills and imitation are the simplest choices that retain the two feedbacks of the model: a positive feedback for adoption, as reliance on AI erodes skill, which makes AI relatively more attractive, and a negative feedback, as crowding lowers the payoff of AI-assisted work when adoption grows. Other choices are possible, and we discuss them in Sec.~\ref{sec:discussion}. Note that two ingredients are not needed to understand the emergence of bistability, crowding $c$, and exploration $\mu$. We therefore first consider $c=0$ and $\mu=0$ in Sections~\ref{sec:return} and~\ref{sec:fate}.
Sec.~\ref{sec:howlong} then considers the role of exploration and Sec.~\ref{sec:crowding} that of crowding. Throughout, we measure time in units of $1/\lambda_{\rm R}$ and use $\beta=0.8$, a strength of selection large enough to depart clearly from weak selection, while keeping Eq.~\eqref{eq:imitation} a probability. The values of the other parameters are given with each figure; our default value, $\lambda=0.2\lambda_{\rm R}$, corresponds to $\rho=0.4$ for $\mu\to0$ and $\rho=0.39$ for $\mu=0.02$.

\section{Population dynamics}\label{sec:exact}

In this section, we derive the equations that describe the time evolution of the population for general values of $c$ and $\mu$. Their key property is that adoption depends on the skills of the population through a single number, the mean skill of non-users, whereas the evolution of this number depends on the whole distribution of skills.

We describe the population by the densities $n_{\rm H}(s,t)$ and $n_{\rm A}(s,t)$ of non-users and AI users with skill $s$, normalised so that $\int(n_{\rm H}+n_{\rm A})\,ds=1$. In a large well-mixed population, they obey the rate equations, also described  in Ref.~\cite{frigerio},
\begin{align}
\partial_t n_{\rm H}&=-\partial_s\big[\lambda(1-s)\,n_{\rm H}\big]-\kHA(s)\,n_{\rm H}+\kAH\,n_{\rm A},\nonumber\\
\partial_t n_{\rm A}&=\partial_s\big[\lambda s\,n_{\rm A}\big]+\kHA(s)\,n_{\rm H}-\kAH\,n_{\rm A}.
\label{eq:rate}
\end{align}
The first terms transport skills along the deterministic flows~\eqref{eq:skill}, and the others describe changes of behaviour, at rates $\kHA(s)$ from H to A and $\kAH$ from A to H. Exploration contributes $\lambda_{\rm R}\mu/2$ to each rate, and imitation a term that depends on the payoff of a randomly chosen peer, through Eq.~\eqref{eq:imitation}. Averaging over peers gives
\begin{align}
\kHA(s)&=\tfrac{\lambda_{\rm R}}{2}\big[\mu+(1-\mu)x\big(1+\beta(b-cx-s)\big)\big],\nonumber\\
\kAH&=\tfrac{\lambda_{\rm R}}{2}\big[\mu+(1-\mu)(1-x)\nonumber\\
&\qquad\times\big(1-\beta(b-cx-m_{\rm H})\big)\big],
\label{eq:rates}
\end{align}
where $x=\int n_{\rm A}\,ds$ is the adoption and $m_{\rm H}$ the mean skill of non-users. Note that a non-user adopts AI more often when their skill is lower, as they gain more by imitating AI users. The rate at which AI users return to working without AI, by contrast, does not depend on their own skill, as their payoff does not. Both rates thus depend on the rest of the population only through two numbers, $x$ and $m_{\rm H}$.

As $\kHA(s)$ is linear in $s$, integrating Eq.~\eqref{eq:rate} over $s$ gives an exact equation for adoption,
\begin{align}
\dot x=\lambda_{\rm R}\Big[&\mu\big(\tfrac12-x\big)\nonumber\\
&+\beta(1-\mu)\,x(1-x)\,\big(b-cx-m_{\rm H}\big)\Big].
\label{eq:adoption}
\end{align}
Adoption thus follows a replicator equation with exploration, in which the differential advantage of AI-assisted work, $b-cx-m_{\rm H}$, is measured against the mean skill of non-users, not of the whole population. Equation~\eqref{eq:adoption} is not closed, however, as $m_{\rm H}$ evolves with the variance of skills among non-users (Sec.~\ref{sec:fate}). With other imitation rules, adoption responds to another statistic of the skills of non-users, e.g.\ a quantile close to their median when agents copy only peers who earn more (Appendix~\ref{sec:app-rate}).

The mean skill $S$ of the population obeys a simpler equation. As switching behaviour does not change the skill of an agent, one finds
\begin{align}
\dot S=\lambda(1-x-S)
\end{align}
for equal rates of learning and erosion, so that $S=1-x$ at stationarity. In other words, in a stationary population, adoption alone fixes the mean skill. Equation~\eqref{eq:adoption} and this relation were obtained at weak selection in Ref.~\cite{frigerio} (their Eqs.~6 and~7), where adoption is also driven by the skill of non-users rather than by the mean skill of the population; with linear imitation, they are exact at any strength of selection.

\section{Thresholds of adoption and return}\label{sec:return}

In this section, we look for the minimal ingredients that produce bistability
and hysteresis. As we show, the individual memory created by skill erosion is
sufficient, and we therefore switch off crowding and exploration, $c=0$ and
$\mu\to0$. Adoption, Eq.~\eqref{eq:adoption}, then obeys
\begin{equation}
\dot x=\lambda_{\rm R}\beta\,x(1-x)\,\big(b-m_{\rm H}\big).
\label{eq:adoptionmin}
\end{equation}
This equation has three kinds of stationary states. In the two absorbing
states, $x=0$, in which all agents are experts, and $x=1$, in which all agents
have lost their skill, no agent has a peer with the other behaviour to imitate,
so that the system becomes frozen. These states can nevertheless be stable
or unstable, depending on whether a few agents with the other behaviour,
introduced e.g.\ by a rare exploration, can spread. Between them lie mixed
states, with $0<x<1$ (Sec.~\ref{sec:branch}). The stability of the absorbing
states, together with the mixed states, determines the long-term dynamics.

\subsection{Mixed stationary states}\label{sec:branch}

We first show that the mixed states form a branch of unstable states that
connects the two absorbing states. By Eq.~\eqref{eq:adoptionmin}, a mixed state
is stationary when non-users are, on average, exactly as productive as
AI-assisted work,
\begin{equation}
b=m_{\rm H}(x).
\label{eq:branch}
\end{equation}
Here $m_{\rm H}(x)$ is the mean skill of non-users in a stationary population with adoption $x$, which depends only on $x$, $\rho$ and $\beta$ (Appendix~\ref{sec:stationary}). In the weak-selection limit, $\beta\to0$,
switching no longer depends on skill and $m_{\rm H}(x)=1-x/(1+\rho)$
(Appendix~\ref{sec:stationary}), which coincides with the stationary skill of autonomous agents in the covariance closure of
Ref.~\cite{frigerio} (their Eq.~19). The mixed states therefore lie on the
straight line
\begin{equation}
x=(1+\rho)(1-b),\qquad \frac{\rho}{1+\rho}<b<1.
\label{eq:branchweak}
\end{equation}

\begin{figure*}[t]
\centering
\includegraphics[width=\linewidth]{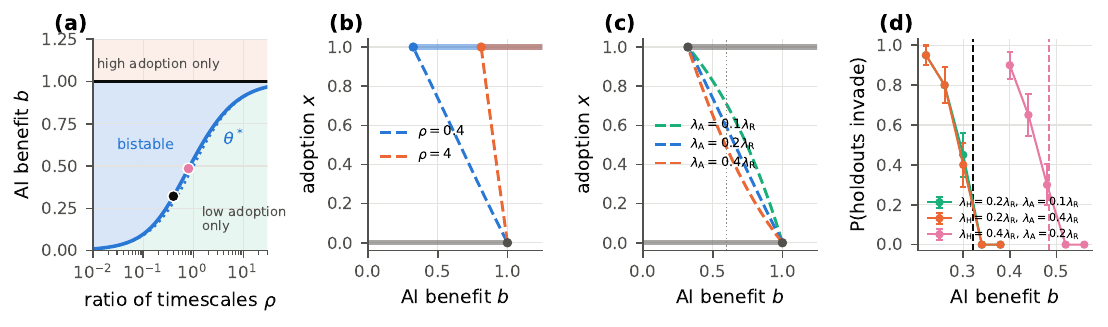}
\caption{\textbf{Thresholds of adoption and return} ($c=0$, $\mu\to0$). (a) Phase diagram in the plane of the ratio of timescales $\rho$ and the benefit $b$ of AI-assisted work. A lone AI user invades a population of experts above $b=1$, and a lone non-user invades a population of AI users below $b=\theta^*(\rho)$, Eq.~\eqref{eq:thetastar}, shown with its weak-selection value $\rho/(1+\rho)$ (dotted). Between the two lines, both absorbing states are stable. The dots mark $\theta^*=0.32$ at $\rho=0.4$ (black) and $\theta^*=0.49$ at $\rho=0.8$ (pink), tested in panel (d). (b) Stationary states against $b$, for $\rho=0.4$ and $\rho=4$: absorbing states without AI ($x=0$, stable for $b<1$) and with AI ($x=1$, stable for $b>\theta^*$), and the unstable mixed branch $b=m_{\rm H}(x)$, Eq.~\eqref{eq:branch} (dashed), which connects them. (c) The same for $\lambda_{\rm H}=0.2\lambda_{\rm R}$ and three rates of erosion $\lambda_{\rm A}$: the mixed branch connects the same thresholds, but lies at lower adoption when skills erode faster; the dotted line marks $b=0.6$. (d) Probability that $20$ holdouts with skill $0$ invade a population of $10^4$ AI users without skill, against $b$ ($\mu=0$, $20$ simulations per point), for two rates of erosion with $\lambda_{\rm H}=0.2\lambda_{\rm R}$, and for $\lambda_{\rm H}=0.4\lambda_{\rm R}$, with binomial standard errors; dashed lines mark $\theta^*$ for $\lambda_{\rm H}=0.2\lambda_{\rm R}$ (black) and $0.4\lambda_{\rm R}$ (pink), i.e.\ $\rho_{\rm H}=0.4$ and $0.8$, the dots of panel (a).}
\label{fig:thresholds}
\end{figure*}

At finite $\beta$, the least skilled non-users are the first to adopt AI, which
raises $m_{\rm H}(x)$ for any $x>0$: the branch ends at a value $m_{\rm H}(1)$ higher than $\rho/(1+\rho)$ for
$x\to1$, which we identify in Sec.~\ref{sec:thresholds} as the threshold of return $\theta^*$.
Note that the branch remains nearly
straight, within $0.005$ for the parameters of Fig.~\ref{fig:thresholds}(b).
In all cases, $m_{\rm H}(x)$ decreases from $1$ for $x\to0$ to $m_{\rm H}(1)$ for
$x\to1$\footnote{We have checked numerically that this decrease is monotonic
for $\rho$ between $0.03$ and $10$.}: the more agents use AI, the more the
remaining non-users are recent returners from AI who have not yet regained
their skill. Along the branch, adoption thus decreases as $b$ increases, from
$x=1$ at $b=m_{\rm H}(1)$ to $x=0$ at $b=1$, and the branch connects the two
absorbing states [Fig.~\ref{fig:thresholds}(b)].

Because $m_{\rm H}(x)$ decreases with $x$, the mixed states are unstable. If
adoption is slightly above that of a mixed state, the non-users, once their
skills have adjusted, tend to earn less than AI-assisted work, and more agents adopt
AI; if it is slightly below, they earn more, and adoption decreases further.
This argument assumes that skills adjust faster than adoption changes;
numerical integration of Eq.~\eqref{eq:rate} confirms the instability. The
mixed states can be understood, in broad terms, as the boundary between the basins of attraction of the
two absorbing states, although, as we show in Sec.~\ref{sec:fate}, adoption
alone does not locate this boundary.

\subsection{Invasion by lone agents}\label{sec:thresholds}

The stability of the absorbing states is determined by whether a lone agent choosing the other behaviour can spread. A lone AI user in a population of experts, whose skills are all equal to $1$, earns $b$. Near $x=0$, $m_{\rm H}=1$ and Eq.~\eqref{eq:adoptionmin} reduces to $\dot x\simeq\lambda_{\rm R}\beta x(b-1)$, so that AI invades when $b>1$: a population of experts adopts AI only when AI-assisted work earns more than expert work, whatever the value of $\rho$.

A lone non-user in a population of AI users, whom we call a holdout, faces a different situation. Their individual memory works against them: they start with almost no skill, lost during their past use of AI, but their skill grows as long as they hold out, as $s(t)=1-e^{-\lambda t}$ after a time $t$ [Eq.~\eqref{eq:skill}], which makes them increasingly attractive to imitate, although they may switch back to AI at any time. Each AI user whom they convert also starts with almost no skill and becomes a new holdout, so that holdouts form a branching process: the population returns to independent work if a holdout converts, on average, more than one AI user before switching back. We now show that this happens if and only if the skill of a holdout, averaged over the time they hold out, exceeds the payoff of AI-assisted work.

Let $\theta=b-c$ be the payoff of AI-assisted work in a population of AI users. For the model parameters considered in this section, $\theta=b$ as $c=0$. A holdout of skill $s$ returns to AI at rate $\kHA(s)=\tfrac{\lambda_{\rm R}}{2}[1+\beta(\theta-s)]$, i.e.\ Eq.~\eqref{eq:rates} with $x\to1$ and $\mu=0$. As their skill increases deterministically, the probability $Q(s)$ that they still hold out when their skill reaches $s$ obeys $dQ/ds=-\kHA(s)\,Q/[\lambda(1-s)]$, which gives $Q(s)=(1-s)^Ae^{-\omega s}$, with
\begin{equation}
A=\frac{\lambda_{\rm R}[1+\beta(\theta-1)]}{2\lambda},\qquad \omega=\frac{\lambda_{\rm R}\beta}{2\lambda}.
\label{eq:holdoutexp}
\end{equation}
These are the exponents $A$ and $\omega$ of the stationary densities of Appendix~\ref{sec:stationary}, evaluated at $x=1$ and $\mu=0$. The time that a holdout spends with skill between $s$ and $s+ds$ is $Q(s)\,ds/[\lambda(1-s)]$, so that their skill, sampled over the time they hold out, has density proportional to $(1-s)^{A-1}e^{-\omega s}$. A holdout of skill $s$ attracts AI users at rate $\tfrac{\lambda_{\rm R}}{2}[1+\beta(s-\theta)]$, as each of the $N$ AI users revises at rate $\lambda_{\rm R}$ and observes the holdout with probability $1/N$. 

Let $T$ be the expected time during which a holdout holds out, and $\bar s$ their skill averaged over this time, i.e.\ the mean of the density above. Both rates are linear in skill, so that, over the time a holdout holds out, their average values are obtained by replacing $s$ by $\bar s$. The expected number of times a holdout switches back is thus $T\tfrac{\lambda_{\rm R}}{2}[1+\beta(\theta-\bar s)]$, which equals $1$, as switching back ends their time as a holdout. The expected number of AI users that they convert is, likewise, $T\tfrac{\lambda_{\rm R}}{2}[1+\beta(\bar s-\theta)]$, and the ratio of the two, $[1+\beta(\bar s-\theta)]/[1+\beta(\theta-\bar s)]$, is the average number of new holdouts that each holdout creates. It exceeds $1$ if and only if $\bar s>\theta$: the population returns to working without AI if the skill of a holdout, averaged over the time they hold out, exceeds the payoff of AI-assisted work. As $\theta$ increases, $A$ increases and $\bar s$ decreases, i.e.\ holdouts return to AI more quickly when AI pays more, so that holdouts invade if and only if $\theta<\theta^*$, where $\theta^*$ is the payoff at which the average skill of a holdout equals the payoff of AI-assisted work, $\theta^*=\bar s(\theta^*)$, with
\begin{equation}
\bar s(\theta)=\frac{\int_0^1 s\,(1-s)^{A-1}e^{-\omega s}\,ds}{\int_0^1(1-s)^{A-1}e^{-\omega s}\,ds},
\label{eq:thetastar}
\end{equation}
which depends on $\theta$ through $A$, Eq.~\eqref{eq:holdoutexp}.

This threshold is also where the mixed branch ends, $m_{\rm H}(1)=\theta^*$. As $x\to1$, the few non-users are holdouts, whose skills follow the density above, so that $m_{\rm H}$ tends to $\bar s$, evaluated at $\theta=b$. The branch condition $b=m_{\rm H}$ then becomes $b=\bar s(b)$, the condition that defines $\theta^*$: the unstable mixed state reaches $x=1$ exactly at $b=\theta^*$, where the state of full adoption loses its stability.

The population is therefore bistable for
\begin{equation}
\theta^*(\rho)<b<1,
\label{eq:holdout}
\end{equation}
whose bounds are shown in Fig.~\ref{fig:thresholds}(a). The threshold $\theta^*$ increases with $\rho$: the more persistent behaviour is compared with the time needed to relearn skills, the more skill a holdout recovers before returning to AI, and the more likely they are to attract imitators. In the weak-selection limit, $\beta\to0$, Eq.~\eqref{eq:holdoutexp} gives $\omega\to0$ and $A\to\lambda_{\rm R}/(2\lambda)=1/\rho$, as $\rho=2\lambda/\lambda_{\rm R}$ for $\mu\to0$. The density of the skill of a holdout then reduces to $(1-s)^{1/\rho-1}$, a $\mathrm{Beta}(1,1/\rho)$ distribution whose mean gives
\begin{equation}
\theta^*=\frac{\rho}{1+\rho}.
\label{eq:thetaweak}
\end{equation}
This is the end point of the weak-selection branch, Eq.~\eqref{eq:branchweak}, and, as for the branch, this simple form holds only for $\beta\to0$. At finite $\beta$, Eq.~\eqref{eq:thetastar} gives a larger threshold, as the least skilled holdouts are the first to switch back: for $\beta=0.8$ and $\rho=0.4$, $\theta^*=0.32$, against $0.29$ at weak selection. Simulations support this threshold [Fig.~\ref{fig:thresholds}(d)]: in a population of $10^4$ AI users without skill, the probability that $20$ holdouts bring adoption below one half, which we use as a criterion of invasion in a finite population, vanishes between $b=0.30$ and $0.34$. As $\theta^*<1$ for any $\rho$, skill erosion alone always produces a bistable window, which narrows as behaviour becomes more persistent but never closes. When behaviour changes much faster than skill ($\rho\to0$), all agents carry nearly the same skill: holdouts switch back to AI before regaining any skill, so that $\theta^*\to0$ and the population is bistable for $0<b<1$, as in a description in which every agent carries the mean skill of the population. In this limit, individual memory disappears, as the skill of an agent no longer depends on their current behaviour, and the model reduces to a description in terms of adoption and mean skill, as in population-level models~\cite{park2026enrichment}; individual memory, which grows with $\rho$, thus eases the return. 

Without erosion, all skills would remain equal to $1$, and the population would adopt AI for $b>1$ and return for $b<1$: the two thresholds would coincide. Skill erosion leaves the threshold of adoption unchanged and lowers the threshold of return from $1$ to $\theta^*$. The two thresholds thus play asymmetric roles. The persistence of behaviour does not change the threshold at which AI invades a population of experts, but it eases the return: the more persistent the behaviours, the more skill the holdouts accumulate, and the more easily the population returns. Skill erosion is thus a cost of returning to working without AI, rather than of adopting it: a holdout starts without skill and earns little until they have regained it.

\subsection{Relearning sets the threshold of return, erosion the unstable state}\label{sec:rates}

So far, skills grow with practice and erode with reliance on AI at the same rate $\lambda$. The two rates may, however, differ: a skill can be slow to acquire and quick to lose, or retained long after practice has stopped. With $\lambda_{\rm H}\neq\lambda_{\rm A}$ in Eq.~\eqref{eq:skill}, the rate equations~\eqref{eq:rate} keep their form, with transport terms $\lambda_{\rm H}(1-s)\,n_{\rm H}$ and $\lambda_{\rm A}s\,n_{\rm A}$.

The two thresholds of Sec.~\ref{sec:thresholds} do not depend on the rate of erosion. A population of experts has skill $1$ whatever the value of $\lambda_{\rm A}$, and adopts AI for $b>1$. A holdout starts without skill and relearns it at rate $\lambda_{\rm H}$, so that Eqs.~\eqref{eq:holdoutexp} and~\eqref{eq:thetastar} hold with $\lambda=\lambda_{\rm H}$: the threshold of return, $\theta^*(\rho_{\rm H})$, depends only on how quickly skills are relearned, e.g.\ $\theta^*=0.32$ and $0.49$ for $\lambda_{\rm H}=0.2\lambda_{\rm R}$ and $0.4\lambda_{\rm R}$. The bistable window $\theta^*<b<1$ thus depends on learning, not on erosion, and the mixed branch connects the same two thresholds whatever the value of $\lambda_{\rm A}$ [Fig.~\ref{fig:thresholds}(c)]. Simulations support this independence [Fig.~\ref{fig:thresholds}(d)]: the probability that holdouts invade is the same for $\lambda_{\rm A}=0.1\lambda_{\rm R}$ and $0.4\lambda_{\rm R}$, and vanishes near $\theta^*=0.32$, whereas for $\lambda_{\rm H}=0.4\lambda_{\rm R}$ it vanishes near $\theta^*=0.49$. This independence also explains why, with exploration, halving or doubling the rate of erosion barely changes the bistable window (Appendix~\ref{sec:app-abm}).

Erosion acts instead inside the bistable window, on the position of the unstable state. Faster erosion makes the remaining non-users less skilled at a given adoption, and thus lowers the mixed branch, while leaving its end points unchanged. At weak selection, for instance, the stationary densities give (Appendix~\ref{sec:stationary})
\begin{equation}
m_{\rm H}(x)=[1+(1-x)/\rho_{\rm A}]/[1+x/\rho_{\rm H}+(1-x)/\rho_{\rm A}],
\end{equation}
 which goes from $1$ to $\rho_{\rm H}/(1+\rho_{\rm H})$ whatever $\rho_{\rm A}$, and is linear in $x$ only for $\rho_{\rm A}=\rho_{\rm H}$. The unstable state thus lies at lower adoption when skills erode faster: for $b=0.6$ and $\lambda_{\rm H}=0.2\lambda_{\rm R}$, its adoption is $0.71$, $0.59$ and $0.48$ for $\lambda_{\rm A}=0.1\lambda_{\rm R}$, $0.2\lambda_{\rm R}$ and $0.4\lambda_{\rm R}$ [Fig.~\ref{fig:thresholds}(c)]. For populations whose skills have adjusted to their behaviour, fewer AI users are then needed to tip them into dependence; away from such distributions, adoption alone does not decide the outcome (Sec.~\ref{sec:fate}). We show in Secs.~\ref{sec:fate} and~\ref{sec:howlong} that erosion also sets how much the expertise of AI users is worth for a return, and how long a temporary advantage of AI must last to lock a population in. Learning and erosion thus play distinct roles: the rate at which skills are relearned decides whether a population can return at all, whereas the rate at which they erode decides how easily it tips into dependence.

\section{What determines the fate of a population?}\label{sec:fate}

\begin{figure*}[t]
\centering
\includegraphics[width=\linewidth]{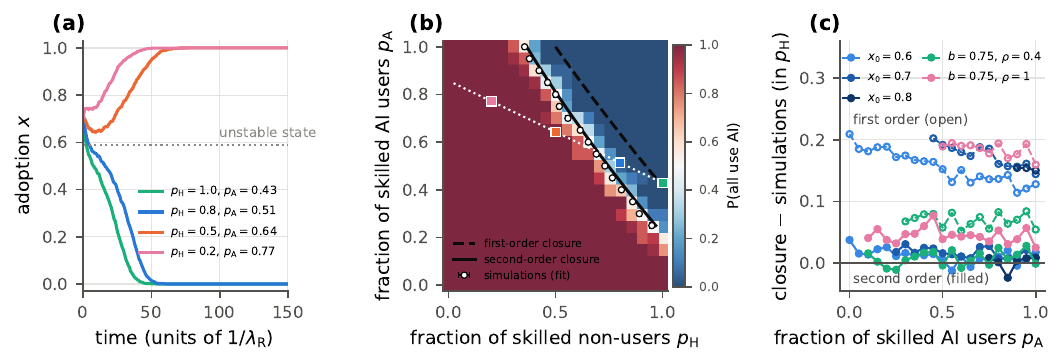}
\caption{\textbf{Same adoption and mean skill, different fates} ($c=0$, $\mu=0$, $b=0.6$, $\rho=0.4$; the unstable stationary state has adoption $0.59$ and mean skill $0.41$). Populations start with adoption $x_0=0.7$; a fraction $p_{\rm H}$ of non-users and a fraction $p_{\rm A}$ of AI users have skill $1$, the others skill $0$. (a) Adoption over time for four initial conditions with the same mean skill, $0.6$ (simulations of $10^4$ agents). (b) Probability that the population ends with everyone using AI, in $20$ simulations of $5000$ agents per point, with the boundary between the two outcomes located by logistic fits (points, with standard errors), the boundaries obtained by integrating the first-order (dashed) and second-order (solid) closures, and the four initial conditions of panel (a) (squares), which lie on a line of constant mean skill (white, dotted). (c) Difference between the boundaries predicted by the first-order (open symbols) and second-order (filled symbols) closures and those measured in simulations, for $b=0.6$, $\rho=0.4$ and three initial adoptions $x_0$, and for $b=0.75$ with $\rho=0.4$ and $\rho=1$ (Appendix~\ref{sec:app-moments}). Boundaries are shown for the values of $p_{\rm A}$ at which they lie within $0\le p_{\rm H}\le1$; for smaller $p_{\rm A}$, and especially at high initial adoption, populations end with everyone using AI even when all non-users are skilled.}
\label{fig:fate}
\end{figure*}

The bistable window of Eq.~\eqref{eq:holdout} raises the question of which state a population reaches from a given initial condition. One might expect adoption, or adoption and mean skill, to answer it. Equation~\eqref{eq:adoption} shows that they cannot, at least initially: as adoption responds to the mean skill of non-users, two populations with the same adoption and mean skill, but in which expertise is held by different groups, can initially move in opposite directions. The fate of a population may thus depend on the individual memories of its members, i.e.\ on how skill, shaped by past AI use, is shared between the two groups.

Figure~\ref{fig:fate} shows that such populations can also end in opposite states. We consider $b=0.6$ and $\rho=0.4$, inside the bistable window, without exploration, so that every population ends in one of the two absorbing states. Initially, a fraction $x_0$ of agents use AI, a fraction $p_{\rm H}$ of non-users and a fraction $p_{\rm A}$ of AI users have skill $1$, and the others skill $0$. The initial mean skills of the two groups are thus $m_{\rm H}=p_{\rm H}$ and $m_{\rm A}=p_{\rm A}$, with variances $p_{\rm H}(1-p_{\rm H})$ and $p_{\rm A}(1-p_{\rm A})$, and the mean skill of the population is $S=(1-x_0)p_{\rm H}+x_0p_{\rm A}$. In Fig.~\ref{fig:fate}(a), all populations start with an adoption of $0.7$ and a mean skill of $0.6$, both larger than those of the unstable stationary state, $0.59$ and $0.41$, but skill is shared differently between the two groups. When a large fraction of non-users is skilled, the population returns to independent work, whereas when skill is mostly held by AI users, it ends with everyone using AI.

Over the whole plane of initial conditions [Fig.~\ref{fig:fate}(b)], the boundary between the two outcomes, where each is reached in half of the simulations, is well described by $p_{\rm H}\simeq1.13-0.80\,p_{\rm A}$. It crosses the lines of constant mean skill, along which the mean skill varies from $0.46$ to $0.81$: mean skill does not decide the outcome. Neither does the mean skill of non-users alone, as the boundary is not vertical: skilled AI users also favour a return, because those who switch back to independent work bring their remaining skill with them. Accounting for the sizes of the two groups, the slope of the boundary implies that a unit of skill held by a non-user has about three times the influence of a unit held by an AI user, whose skill keeps eroding until they return. Indeed, a unit of skill raises $p_{\rm H}$ by $1/(1-x_0)$ if it is held by non-users, and $p_{\rm A}$ by $1/x_0$ if it is held by AI users; along the boundary, $dp_{\rm H}=-0.80\,dp_{\rm A}$, so that the ratio of influences is $x_0/[0.80(1-x_0)]\simeq2.9$. In the reduced description of Ref.~\cite{frigerio}, the basins of the two states are separated by a curve in the plane of adoption and mean skill; here, mean skill is not sufficient, as the outcome depends on who holds the skill.

To identify which statistics of the distribution matter, we derive equations for its moments. The aim is not to solve the dynamics exactly, but to find a minimal description that predicts the outcome. We denote by $M_{\rm H}^{(k)}=\int s^k n_{\rm H}\,ds$ and $M_{\rm A}^{(k)}=\int s^k n_{\rm A}\,ds$ the moments of the skills of non-users and AI users, so that $M_{\rm H}^{(0)}=1-x$ and $M_{\rm A}^{(0)}=x$, and by $m_{\rm H}$, $m_{\rm A}$, $V_{\rm H}$ and $V_{\rm A}$ the means and variances of the two groups. Multiplying Eq.~\eqref{eq:rate} by $s^k$ and integrating over skill gives
\begin{align}
\frac{dM_{\rm H}^{(k)}}{dt}&=k\lambda\big(M_{\rm H}^{(k-1)}-M_{\rm H}^{(k)}\big)-a_0M_{\rm H}^{(k)}+a_1M_{\rm H}^{(k+1)}\nonumber\\
&\quad+\kAH M_{\rm A}^{(k)},\nonumber\\
\frac{dM_{\rm A}^{(k)}}{dt}&=-k\lambda M_{\rm A}^{(k)}+a_0M_{\rm H}^{(k)}-a_1M_{\rm H}^{(k+1)}-\kAH M_{\rm A}^{(k)}.
\label{eq:moments}
\end{align}
These equations hold for any distribution of skills, including the initial conditions of Fig.~\ref{fig:fate}, in which all skills are $0$ or $1$. For $k=0$, they reduce to the adoption equation~\eqref{eq:adoption}. The moment of order $k$ is coupled to the moment of order $k+1$ through the non-users, because only their switching rate depends on their own skill. The hierarchy is therefore not closed: the equation for the mean involves the variance, the equation for the variance involves the third moment, and so on. 

For $k=1$, the first equation reads
\begin{align}
\frac{d}{dt}\big[(1-x)m_{\rm H}\big]=\;&\lambda(1-x)(1-m_{\rm H})\nonumber\\
&-\kHA(m_{\rm H})\,(1-x)\,m_{\rm H}\nonumber\\
&+a_1(1-x)V_{\rm H}+\kAH\,x\,m_{\rm A}.
\label{eq:mHdyn}
\end{align}
The four terms have a simple interpretation. Non-users gain skill through practice. They lose skill as they switch to AI, at a rate evaluated at their mean skill. The third term corrects this loss: as the least skilled non-users are the first to adopt AI, departures remove less skill than they would if departing non-users had the mean skill, by an amount proportional to the variance of skills among non-users. Finally, AI users who return to independent work bring their remaining skill with them. Adoption responds to $m_{\rm H}$ only, but $m_{\rm H}$ itself depends on $m_{\rm A}$ and on $V_{\rm H}$, which explains the two features of Fig.~\ref{fig:fate}(b): skilled AI users matter through the last term, and the spread of skills among non-users through the third.

Equations~\eqref{eq:moments} can be closed at two levels. Neglecting the variance of skills among non-users, $V_{\rm H}=0$, closes them at first order, in three variables, $x$, $m_{\rm H}$ and $m_{\rm A}$, a description comparable to the covariance closure of Ref.~\cite{frigerio}. Keeping the second moments closes them at second order, in five variables, once the third moment of non-users is expressed in terms of their mean and variance. We assume that the skills of non-users follow a Beta distribution with the same mean and variance, a natural choice for a variable confined to $[0,1]$, and the form taken, up to an exponential tilt, by the stationary distributions of Appendix~\ref{sec:stationary}. Writing $\kappa=m_{\rm H}(1-m_{\rm H})/V_{\rm H}-1$ for the sum of its two parameters, the mean of $s^3$ among non-users is then
\begin{equation}
\langle s^3\rangle_{\rm H}=\frac{m_{\rm H}\,(\kappa m_{\rm H}+1)(\kappa m_{\rm H}+2)}{(\kappa+1)(\kappa+2)},
\label{eq:betaclosure}
\end{equation}
so that $M_{\rm H}^{(3)}=(1-x)\langle s^3\rangle_{\rm H}$. In particular, the closure is exact for the initial conditions of Fig.~\ref{fig:fate}: when skills are $0$ or $1$, the variance takes its largest value, $\kappa=0$, and Eq.~\eqref{eq:betaclosure} gives $\langle s^3\rangle_{\rm H}=m_{\rm H}$, as it should.

This closure is a moment-matching approximation rather than a controlled expansion: the distributions generated by the dynamics need not be Beta distributions, and the similarity with the stationary densities motivates the choice but does not guarantee its accuracy, which we assess against simulations. The results depend little on the choice of closure for the third moment (Appendix~\ref{sec:app-moments}). The closures and the simulations give the boundary in different ways. The closures are deterministic: we integrate the closed equations from the moments of the initial condition until adoption reaches $0$ or $1$, which gives a sharp boundary with no fitted parameter. In simulations, the outcome of each run is random, and we estimate the boundary for each $p_{\rm A}$ by fitting a logistic function of $p_{\rm H}$ to the outcomes of the runs, with a standard error of about $0.01$ [points in Fig.~\ref{fig:fate}(b)].

The first-order closure misplaces the boundary by $0.15$ to $0.17$ in $p_{\rm H}$ and predicts lock-in too often, whereas the second-order closure locates it within $0.02$ on average, for initial adoptions of $0.6$, $0.7$ and $0.8$ [Fig.~\ref{fig:fate}(c)]. For $b=0.75$, the errors of the second-order closure are $0.01$ for $\rho=0.4$ and $0.05$ for $\rho=1$, against $0.07$ and $0.18$ for the first-order closure: the second-order closure is thus less accurate when behaviour is persistent and skills are strongly polarised within each group (Appendix~\ref{sec:app-moments}). The simulated boundary hardly depends on the size of the population: when the number of agents increases tenfold, it moves by less than $0.01$ (Appendix~\ref{sec:app-moments}).

\begin{table}[t]
\caption{\label{tab:predict}Fraction of runs (in \%) whose outcome is correctly predicted from the initial condition, for $c=0$, $\mu=0$ and the five settings of Fig.~\ref{fig:fate}(c): (i--iii) $b=0.6$, $\rho=0.4$ and $x_0=0.6$, $0.7$ and $0.8$; (iv) $b=0.75$, $\rho=0.4$, $x_0=0.42$; (v) $b=0.75$, $\rho=1$, $x_0=0.57$. Each setting includes all the runs of its map ($8820$ for (ii), $4410$ for the others). The first row always predicts the outcome most frequent overall, and the second uses the best threshold on the initial mean skill $S$, fitted on half of the runs at each initial condition and scored on the other half. The closures involve no fitting. The last row predicts, at each initial condition, the outcome most frequent in the other half of the runs, an empirical benchmark limited by finite-size fluctuations.}
\begin{ruledtabular}
\begin{tabular}{lccccc}
predictor & (i) & (ii) & (iii) & (iv) & (v)\\
\hline
majority outcome & 55.0 & 71.4 & 94.4 & 82.2 & 77.5\\
mean skill $S$ & 80.8 & 85.6 & 96.6 & 94.1 & 86.7\\
closure, 1st order & 85.1 & 87.8 & 96.4 & 94.1 & 84.3\\
closure, 2nd order & 96.7 & 96.7 & 98.3 & 97.1 & 93.0\\
benchmark & 96.2 & 96.5 & 97.9 & 96.5 & 93.4\\
\end{tabular}
\end{ruledtabular}
\end{table}

Table~\ref{tab:predict} summarises how well each description predicts the outcome of individual runs. The initial mean skill misclassifies up to $16$ percentage points more runs than the second-order closure, and the first-order closure improves on it by at most about $4$ percentage points. The second-order closure, which involves no fitting, is as accurate as the empirical benchmark, except for $\rho=1$, where it is $0.4$ percentage points lower. The first-order closure fails even for homogeneous initial skills (Appendix~\ref{sec:app-moments}), because the dynamics itself creates a variance of skills among non-users, whose selective departure then raises their mean skill.

The rate of erosion changes how much the expertise of AI users is worth [Fig.~\ref{fig:fateapp}(f)]: for $\lambda_{\rm A}=0.1\lambda_{\rm R}$, $0.2\lambda_{\rm R}$ and $0.4\lambda_{\rm R}$, skill held by an AI user has about $0.4$, $0.3$ and $0.2$ times the influence of skill held by a non-user, as it erodes more quickly before they return, and the second-order closure, with $k\lambda_{\rm H}$ and $k\lambda_{\rm A}$ in the two equations~\eqref{eq:moments}, locates these boundaries within $0.02$ on average. The same picture holds when the initial skills within each group follow Beta distributions rather than being $0$ or $1$ (Appendix~\ref{sec:app-moments}).

\section{Exploration and temporary advantages}\label{sec:howlong}\label{sec:exploration}

\begin{figure*}[t]
\centering
\includegraphics[width=0.75\linewidth]{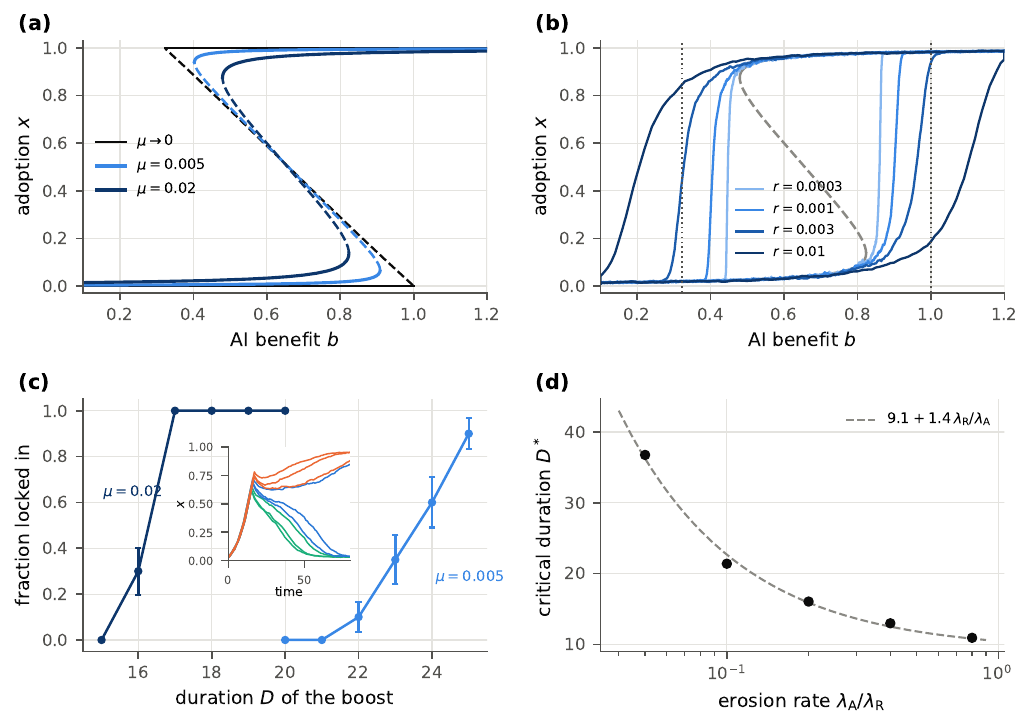}
\caption{\textbf{Exploration and temporary advantages} ($c=0$; $\lambda=0.2\lambda_{\rm R}$ unless stated otherwise). (a) Stationary adoption against $b$ for $\mu\to0$ (black: absorbing states and unstable mixed branch), $\mu=0.005$ and $\mu=0.02$ (exact solution, dashed on the unstable branch). (b) Adoption against $b$ when $b$ rises from $0.1$ to $1.2$ at a rate $r$, remains at $1.2$ for $300$ time units and falls back at the same rate ($\mu=0.02$, simulations of $10^4$ agents), with the exact stationary solution (grey) and the thresholds $b=\theta^*$ and $b=1$ for $\mu\to0$ (dotted). (c) Fraction of $20$ simulations of $10^4$ agents that remain at high adoption after $b$ is raised from $0.6$ to $1.2$ for a duration $D$ and then restored to $0.6$, with binomial standard errors. Inset: adoption over time in three runs for each of $D=15$, $16$ and $17$ ($\mu=0.02$). (d) Critical duration $D^*$ of such a boost against the rate of erosion $\lambda_{\rm A}$, for $\mu=0.02$ and $\lambda_{\rm H}=0.2\lambda_{\rm R}$, with the fit $D^*\simeq9.1+1.4\,\lambda_{\rm R}/\lambda_{\rm A}$ (dashed).}
\label{fig:dynamics}
\end{figure*}

For the purpose of this section, it is necessary to include exploration, $\mu>0$. Exploration ensures that a population in which all agents, or none, use AI can still change behaviour, as explorers provide peers to imitate. Without it, a population of experts would never adopt AI, however large its benefit, as it would have no AI user to imitate. Exploration also represents processes that the imitation rule leaves out, e.g.\ newcomers, curiosity or a tool becoming temporarily unavailable. For $\mu>0$, the homogeneous states become states of low and high adoption with a minority of explorers, described by the stationary densities of Appendix~\ref{sec:stationary}. Each of these stable states now disappears at a fold, i.e.\ a value of $b$ at which it merges with the unstable branch [Fig.~\ref{fig:dynamics}(a)]. The folds lie inside the invasion thresholds of Eq.~\eqref{eq:holdout}, because exploration continually injects agents of the minority behaviour, so that each stable state loses its stability before a lone invader would succeed on its own. For $\lambda=0.2\lambda_{\rm R}$, the bistable window shrinks from $0.32<b<1$ for $\mu\to0$ to $0.40<b<0.91$ for $\mu=0.005$ and $0.48<b<0.82$ for $\mu=0.02$. Exploration also makes the threshold of adoption depend on $\rho$, as agents who have recently stopped using AI lower the mean skill of non-users, the more so the smaller $\rho$: for $\mu=0.02$, this threshold rises from $0.82$ for $\rho=0.39$ to $0.89$ for $\rho=2$. Unlike without exploration (Sec.~\ref{sec:return}), the bistable window then closes at a finite persistence, near $\rho\approx8.5$ for $\mu=0.02$.

Exploration is required to explore how the population reacts to changes of the 
benefit of AI  over time, as tools improve or become less expensive for instance. We consider two scenarios. In the first, the benefit increases and then decreases at a constant rate $r$ [Fig.~\ref{fig:dynamics}(b)]. As AI improves, adoption remains low until the benefit crosses the upper fold, after which it rises abruptly. The faster the increase, the larger the benefit at which adoption takes off, because the minority of AI users maintained by exploration needs time to grow; the return to independent work is likewise delayed on the way down. For $\mu=0.02$, adoption crosses $\tfrac12$ at $b=0.86$ on the way up and $0.45$ on the way down for $r=3\times10^{-4}$, close to the folds at $0.82$ and $0.48$, but at $1.11$ and $0.19$ for $r=10^{-2}$. Exploration thus sets the size of the seed from which change grows, and hence the delay of the response. For $\mu=0.005$, transitions on slow ramps occur closer to the thresholds without exploration, $b=1$ and $\theta^*$, at $b=0.95$ and $0.36$, but the response is so slow that, for $r=10^{-2}$, the population changes state only after the benefit has stopped changing. For illustration, if agents reconsider their use of AI about once a week, so that skills adapt over about five weeks, increasing $b$ from $0.1$ to $1.2$ takes about seven years for $r=3\times10^{-3}$ and two years for $r=10^{-2}$.

In the second scenario, a temporary increase of the benefit, e.g.\ a period of low prices, lasts for a duration $D$ [Fig.~\ref{fig:dynamics}(c)]. The benefit is raised from $0.6$, where the population is in the state of low adoption, to $1.2$, above both thresholds of adoption, and then restored to $0.6$. As $b=0.6$ lies inside the bistable window, both states are stable there, and lock-in is expected if the increase moves the population into the basin of high adoption; the question is how long the increase must last. Short increases have little lasting effect, whereas sufficiently long ones leave the population in the state of high adoption, although the benefit has returned to its initial value: the increase leaves a lasting trace in individual memories, as the skills lost during the increase make independent work less attractive, a possibility noted qualitatively in Ref.~\cite{frigerio}. We define the critical duration $D^*$ as the duration at which the fraction of runs that remain at high adoption, fitted by a logistic function of $D$, reaches one half, for the population size and observation time used here. For $\mu=0.02$, no run remains at high adoption for $D=15$, $30\%$ do for $D=16$, and all do from $D=17$, i.e.\ after about three times the timescale $1/\lambda$ of skill adaptation, or four months with the illustrative time unit above. For $\mu=0.005$, the crossover occurs later, near $D\approx24$, and is broader, as the smaller seed of AI users makes the time at which adoption takes off longer and more variable.

The critical duration also depends on how quickly skills erode [Fig.~\ref{fig:dynamics}(d)]: for $\mu=0.02$ and $\lambda_{\rm H}=0.2\lambda_{\rm R}$, it decreases from $D^*\simeq37$ for $\lambda_{\rm A}=0.05\lambda_{\rm R}$ to $21$, $16$, $13$ and $11$ for $\lambda_{\rm A}=0.1\lambda_{\rm R}$, $0.2\lambda_{\rm R}$, $0.4\lambda_{\rm R}$ and $0.8\lambda_{\rm R}$, approximately as $9.1+1.4\,\lambda_{\rm R}/\lambda_{\rm A}$, an empirical fit over the range investigated. This dependence is consistent with two contributions, a delay associated with the growth of the initial minority of AI users and a skill-erosion timescale proportional to $1/\lambda_{\rm A}$, although the fit alone does not separate them. Exploration thus sets both how quickly and how predictably a population responds to changes in the benefit of AI, and erosion how long an advantage must last to lock it in.

\section{Crowding}\label{sec:crowding}

\begin{figure*}[t]
\centering
\includegraphics[width=0.85\linewidth]{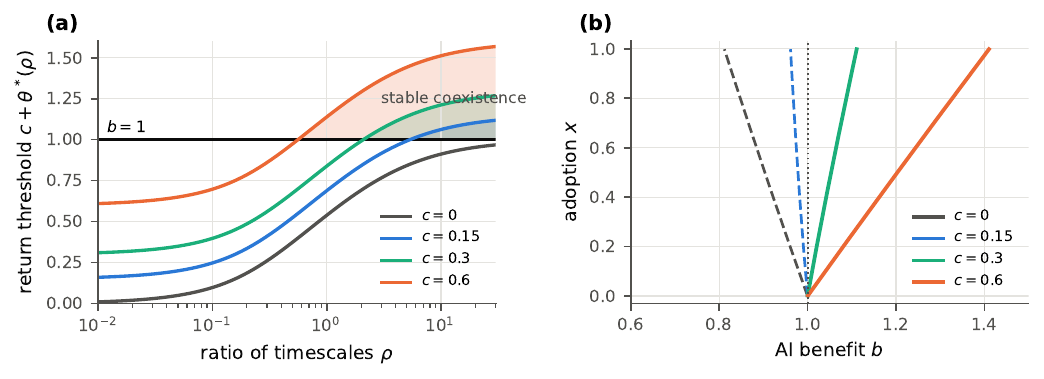}
\caption{\textbf{Crowding} ($\mu\to0$). (a) Threshold of return $b=c+\theta^*(\rho)$ against $\rho$ for four values of $c$, and threshold of adoption $b=1$. Where the threshold of return lies below $b=1$, the population is bistable; where it lies above (shaded), neither absorbing state is stable and the population settles at partial adoption. (b) Stationary adoption against $b$ for $\rho=4$: mixed branch $b=m_{\rm H}(x)+cx$, Eq.~\eqref{eq:branchc}, unstable (dashed) for $c=0$ and $0.15$, and stable (solid) for $c=0.3$ and $0.6$.}
\label{fig:crowding}
\end{figure*}

We finally add the effect of crowding, $c>0$, which lowers the payoff of AI-assisted work as adoption grows, and thus acts against the feedback of skill erosion. 
Without exploration, its effect can be computed exactly. As $m_{\rm H}(x)$ does not depend on $c$ (Appendix~\ref{sec:stationary}), Eq.~\eqref{eq:branch} becomes
\begin{equation}
b=m_{\rm H}(x)+c\,x .
\label{eq:branchc}
\end{equation}
Crowding tilts the mixed branch [Fig.~\ref{fig:crowding}(b)] and shifts the threshold of return to $b=c+\theta^*(\rho)$, as the payoff of AI-assisted work in a population of AI users becomes $\theta=b-c$ in Eq.~\eqref{eq:thetastar}, while leaving the threshold of adoption at $b=1$, as a lone AI user does not crowd anybody. The population is bistable for $c+\theta^*(\rho)<b<1$. When $c+\theta^*(\rho)>1$, the two invasion conditions are satisfied together: a lone AI user invades experts for $b>1$, and a lone holdout invades AI users for $b<c+\theta^*$. Neither absorbing state is then stable and, in the parameter regimes investigated, the mixed branch is stable, so that adoption increases continuously with $b$, from $0$ at $b=1$ to $1$ at $b=c+\theta^*$. Crowding thus replaces hysteresis by a stable coexistence of non-users and AI users, beyond a persistence $\rho_c$ at which $c+\theta^*(\rho_c)=1$, i.e.\ $\rho_c=5.35$, $2.09$ and $0.56$ for $c=0.15$, $0.3$ and $0.6$ [Fig.~\ref{fig:crowding}(a)]. In this coexistence, skills can be strongly polarised between the two groups, a structure that adoption and mean skill do not reveal~\cite{frigerio}. In the weak-selection limit, $\theta^*=\rho/(1+\rho)$ and the population is bistable for $c<1/(1+\rho)$. This coincides with the effective-feedback criterion of Ref.~\cite{frigerio}, $\Gamma_{\rm eff}>8m$: setting their benefit of autonomous work to $1$, removing their skill floor and using our convention for the timescale ratio gives $\Gamma_{\rm eff}=1/(1+\rho)-c$, and the criterion reduces to $\Gamma_{\rm eff}>0$ without exploration ($m=0$). The holdout argument gives this criterion a microscopic interpretation and extends it to finite selection.

With both crowding and exploration, the bistable region closes at a smaller persistence than with either ingredient alone, near $\rho\approx2.7$ for $c=0.15$ and $\mu=0.02$; for $\rho=0.39$, it extends from $b=0.610$ to $0.845$, and simulations of $10^4$ agents reproduce these folds within $0.015$ (Appendix~\ref{sec:app-abm}).

\section{Discussion}\label{sec:discussion}

Technologies may improve immediate performance but also erode the skills needed to work without them, as documented for automation and navigation~\cite{bainbridge1983,dahmani2020} and more recently for AI~\cite{budzyn2025}. We have investigated how, when individuals imitate the behaviour of more successful peers, this erosion turns adoption into a collective process with memory. Skill erosion  makes adoption of AI bistable, with asymmetric thresholds:  a population of experts adopts AI only when AI-assisted work earns more than expert work, whereas a population relying on AI returns only if holdouts relearn enough skill before switching back. Skill erosion therefore acts as a switching cost, as in economics, where switching costs usually arise from investments in equipment or training~\cite{klemperer}. But here, in contrast, adopting AI has no cost, whereas returning requires rebuilding the skill lost while using it, a cost that grows with the duration of AI use.

Because the skill of each agent records their own history, adoption and average skill do not determine the state of a population. Populations with the same adoption and mean skill can reach full adoption or full return, depending on whether expertise is held by non-users or by AI users, and predicting the outcome requires following not only the mean skills of the two groups but also the variance of skills among non-users, which selective switching creates. Individual memory also lets a temporary advantage of AI lock a population in, if it lasts longer than a critical duration that decreases as skills erode faster, while exploration sets how quickly and how predictably the population responds. 

These results suggest  ways of limiting dependence on AI. For instance, supporting those who return to independent work, e.g.\ via training, increases the rate at which skills are relearned and is expected to change the threshold of return. In contrast, tools that keep their users engaged slow the erosion of skills~\cite{gajos,bastani}, and our results suggest that they would leave the thresholds unchanged, but lengthen the advantage needed to lock a population in. The model also makes predictions that could be tested empirically. The ability of a population to return to independent work should depend on the skill of non-users and of recent adopters, rather than on adoption or average skill, and a temporary advantage of AI should have lasting effects only if it lasts longer than a critical duration. Studies of earlier technologies often measure reduced learning or impaired performance during use~\cite{sparrow2011}, and more rarely the erosion of skills that people already had~\cite{dahmani2020,budzyn2025}. Our results suggest that both should be measured, as relearning and erosion play different roles in whether a population can return.

The conclusions of this paper rest on simple modelling choices: a well-mixed population of identical agents,  imitation of realised payoffs, and no calibration against data. The qualitative effects nonetheless persist under other imitation rules (Appendix~\ref{sec:app-abm}), but the precise thresholds do not. Natural extensions would include imposing a social network structure to the agents, for instance made of communities, heterogeneous agents and tasks, forward-looking decisions, and calibration against longitudinal data on AI use and unassisted performance. 

\section*{Code availability}
The code that reproduces all the figures, including the exact stationary solution, the lone-holdout threshold, the moment closures and the agent-based simulations, is available at \url{https://github.com/rlambiot/modelling-lock-in-ai-adoption-skill-erosion}.

\begin{acknowledgments}
I warmly thank R. Baeza-Yates, D. Pedreschi, L. Pappalardo, and N. Oliver for discussions on human--AI coevolution.
I used Claude Opus 5.5 (Anthropic) as an AI assistant in this work. It was used
to write and debug the code for the simulations, solvers and figures, to check derivations and proofs, to discuss the interpretation of the results, and to edit the text. I reviewed and edited all stages of the work and take full responsibility for its content. In line with this paper, I tried to use AI in a way that preserved my own independent skills.
I acknowledge support from the EPSRC grants EP/V013068/1, EP/V03474X/1 and EP/Y028872/1.
\end{acknowledgments}

\appendix

\section{Imitation rules}\label{sec:app-rate}

\paragraph{Linear imitation at finite selection.} For the Fermi rule, $P_{\rm F}(i\leftarrow j)=[1+\exp(-\gamma(\pi_j-\pi_i))]^{-1}$, the first-order expansion at weak selection is $\tfrac12+\tfrac{\gamma}{4}(\pi_j-\pi_i)$, and Eq.~\eqref{eq:imitation} is this linear rule with $\gamma=2\beta$. We use Eq.~\eqref{eq:imitation} at finite $\beta$, rather than assuming weak selection, which only requires that it define a probability: $\beta|\pi_j-\pi_i|\le1$. In all the simulations and stationary solutions that we report, the payoff of AI-assisted work, $b-cx$, lies between $0$ and $1.2$; along the mixed stationary branches without exploration, it equals the mean skill of non-users and thus does not exceed $1$, even when $b>1$. Hence $|\pi_j-\pi_i|\le1.2$, $\beta|\pi_j-\pi_i|\le0.96$, and $P$ remains between $0.02$ and $0.98$.

\paragraph{Other imitation rules.} Let $g(\Delta)$ be the probability of copying a peer, where $\Delta$ is the peer's payoff minus one's own. For the Fermi rule, $g(\Delta)=[1+\exp(-\gamma\Delta)]^{-1}$, with selection intensity $\gamma$. When $g(\Delta)+g(-\Delta)=1$, and writing $\theta=b-cx$, adoption obeys
\begin{align}
\dot x=\lambda_{\rm R}\Big[&\mu\big(\tfrac12-x\big)\nonumber\\
&+(1-\mu)x(1-x)\big\langle 2g(\theta-s)-1\big\rangle_{\rm H}\Big],
\label{eq:adoptiongen}
\end{align}
where $\langle\cdot\rangle_{\rm H}$ denotes an average over non-users. Thus, the imitation rule determines which aspect of their skill distribution drives adoption. For linear imitation, it is their mean skill, Eq.~\eqref{eq:adoption}. When agents copy only peers who earn more, adoption instead depends on the fraction of non-users whose skill lies above or below $\theta$: when exploration is weak, adoption is stationary when $\theta$ is close to the median skill of non-users. For the Fermi rule, the switching rates are no longer affine in skill, but Eq.~\eqref{eq:zeroflux} still gives the stationary densities, by quadrature.

\section{Exact stationary distributions of skill}\label{sec:stationary}\label{sec:app-exact}

The folds of the stationary curve with exploration, the mixed branch without exploration and the distributions of skill all follow from the stationary solution of Eq.~\eqref{eq:rate}, which is exact for linear imitation. Non-users drift towards high skill and AI users towards low skill, while changes of behaviour move agents between these two flows. At stationarity, the upward flow of non-users through any level of skill must balance the downward flow of AI users: adding the two equations shows that the flux of probability along $s$, $\lambda(1-s)\,n_{\rm H}-\lambda s\,n_{\rm A}$, is uniform, and it vanishes as no agent leaves $[0,1]$. The upward flux of non-users through each level of skill, $w(s)=\lambda(1-s)\,n_{\rm H}(s)=\lambda s\,n_{\rm A}(s)$, then obeys the first equation of~\eqref{eq:rate} in the form
\begin{equation}
\frac{d\ln w}{ds}=-\frac{\kHA(s)}{\lambda(1-s)}+\frac{\kAH}{\lambda s}.
\label{eq:zeroflux}
\end{equation}
Writing $\kHA(s)=\kHA(1)+a_1(1-s)$ and integrating gives $w\propto(1-s)^A s^B e^{-\omega s}$, and dividing by the skill velocities gives the stationary densities
\begin{align}
n_{\rm H}(s)&=Z\,(1-s)^{A-1}s^{B}e^{-\omega s},\nonumber\\
n_{\rm A}(s)&=Z\,(1-s)^{A}s^{B-1}e^{-\omega s},
\label{eq:tiltedbeta}
\end{align}
with $A=\kHA(1)/\lambda$, $B=\kAH/\lambda$ and $\omega=a_1/\lambda$. The total density of skills, $n_{\rm H}+n_{\rm A}=Z\,s^{B-1}(1-s)^{A-1}e^{-\omega s}$, is thus an exponentially tilted Beta distribution, in which the exponential factor accounts for skill-dependent switching. The zero-flux construction was also introduced in Ref.~\cite{frigerio} (Supplementary Information, Sec.~S10), where the tilt appears at first order in the strength of selection for Fermi imitation; with linear imitation, the switching rates are exactly affine in skill, and Eq.~\eqref{eq:tiltedbeta} is exact at any strength of selection for which Eq.~\eqref{eq:imitation} defines a probability.

The rates entering Eq.~\eqref{eq:tiltedbeta} depend themselves on $x$ and $m_{\rm H}$, which must agree with the values obtained from the densities. For a chosen $0<x<1$, the stationary adoption equation~\eqref{eq:adoption} fixes the payoff advantage,
\begin{equation}
b-cx-m_{\rm H}=\frac{\mu\,(x-\tfrac12)}{\beta(1-\mu)\,x(1-x)},
\label{eq:advantage}
\end{equation}
which determines $B$ and $\omega$. The remaining exponent $A$ is fixed by requiring the densities to reproduce the chosen adoption, $x\,I_{\rm H}=(1-x)\,I_{\rm A}$, where $I_{\rm H}$ and $I_{\rm A}$ are the integrals of the unnormalised densities, given in closed form by Kummer functions below. The densities then give $m_{\rm H}$, and Eq.~\eqref{eq:advantage} gives $b$. Varying $x$ thus traces the curve of stationary states, including its unstable parts, and locates its folds, but does not establish stability.

For $\mu>0$, $A$ and $B$ are positive and the densities are integrable. For $\mu=0$, the right-hand side of Eq.~\eqref{eq:advantage} vanishes, so that $b-cx=m_{\rm H}$, and the switching rates, and hence the densities, depend only on $x$, $\rho$ and $\beta$: the mean skill of non-users on the mixed branch, $m_{\rm H}(x)$, depends neither on $b$ nor on $c$. These densities describe the mixed stationary states of Sec.~\ref{sec:branch}, whose mean skill of non-users gives the branch~\eqref{eq:branch}; the holdout density of Sec.~\ref{sec:thresholds} is their limit for $B\to0$.

In the weak-selection limit, $\beta\to0$, the switching rates no longer depend on skill and $\omega=0$. On the mixed branch, $\mu=0$ and $b=m_{\rm H}$, so that $A=x/\rho$ and $B=(1-x)/\rho$, and the skills of non-users follow a $\mathrm{Beta}(B+1,A)$ distribution, with mean
\begin{equation}
m_{\rm H}(x)=\frac{B+1}{A+B+1}=1-\frac{x}{1+\rho}.
\label{eq:mHweak}
\end{equation}

With unequal rates of learning and erosion, $w=\lambda_{\rm H}(1-s)\,n_{\rm H}=\lambda_{\rm A}s\,n_{\rm A}$, and the same construction gives the densities~\eqref{eq:tiltedbeta}, divided by $\lambda_{\rm H}$ and $\lambda_{\rm A}$, respectively, with $A=\kHA(1)/\lambda_{\rm H}$, $B=\kAH/\lambda_{\rm A}$ and $\omega=a_1/\lambda_{\rm H}$. At weak selection, on the mixed branch, $A=x/\rho_{\rm H}$ and $B=(1-x)/\rho_{\rm A}$, and the mean skill of non-users, $(B+1)/(A+B+1)$, becomes $m_{\rm H}(x)=[1+(1-x)/\rho_{\rm A}]/[1+x/\rho_{\rm H}+(1-x)/\rho_{\rm A}]$.

To normalise the densities~\eqref{eq:tiltedbeta} and calculate their moments, we use
$\int_0^1s^{p-1}(1-s)^{q-1}e^{-\omega s}\,ds=\mathrm B(p,q)\,{}_1F_1(p;p+q;-\omega)$, where $\mathrm B$ and ${}_1F_1$ are the Beta and Kummer functions. The integrals of the two unnormalised densities are
\begin{align}
I_{\rm H}&=\mathrm B(B+1,A)\;{}_1F_1(B+1;A+B+1;-\omega),\nonumber\\
I_{\rm A}&=\mathrm B(B,A+1)\;{}_1F_1(B;A+B+1;-\omega).
\label{eq:kummer}
\end{align}
Thus $Z=1/(I_{\rm H}+I_{\rm A})$, $x=I_{\rm A}/(I_{\rm H}+I_{\rm A})$, and the mean skill of non-users is
\begin{equation}
m_{\rm H}=\frac{\mathrm B(B+2,A)\;{}_1F_1(B+2;A+B+2;-\omega)}{I_{\rm H}}.
\label{eq:mHkummer}
\end{equation}
For equal rates, the mean skill of AI users follows from the stationary identity $S=1-x$: $m_{\rm A}=(1-x)(1-m_{\rm H})/x$. The exponent $A$ is both defined by $A=\kHA(1)/\lambda$ and fixed by the condition on adoption. The two agree, as the stationary flux balance and the adoption equation require the same total rate of switching from H to A, with the same dependence on skill. For $\mu=0$, the homogeneous states without and with AI correspond to point masses at $s=1$ and $s=0$, respectively.

\section{Fate of populations: other parameters, initial distributions and rates}\label{sec:app-moments}

\begin{figure*}[t]
\centering
\includegraphics[width=\linewidth]{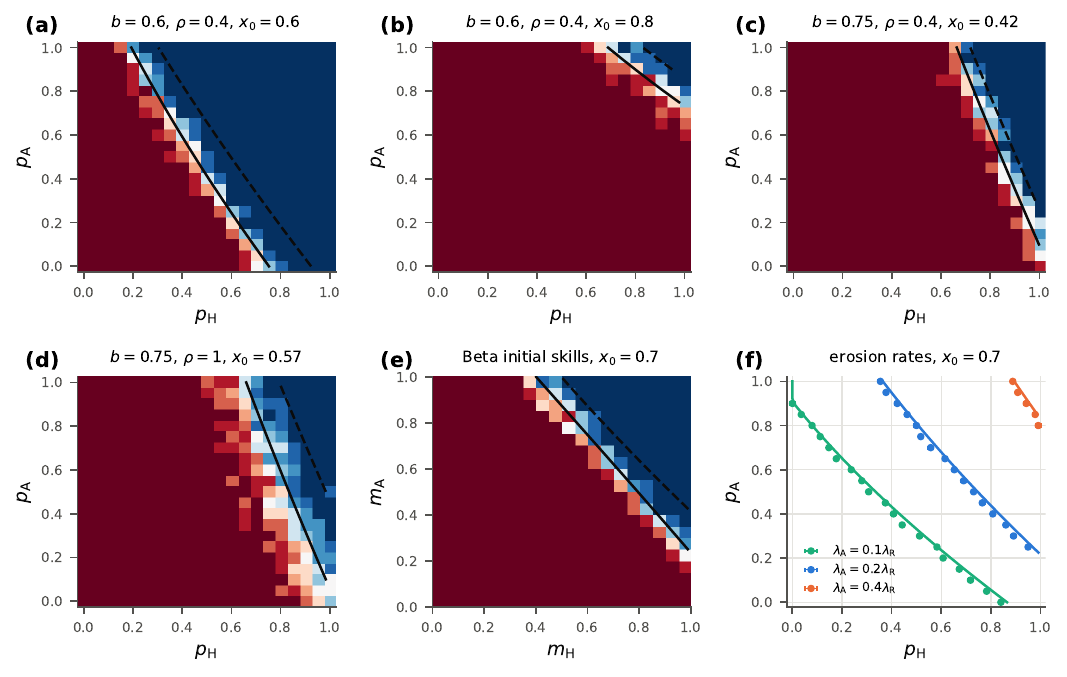}
\caption{\textbf{Fate of populations at other parameters, initial distributions and rates} ($c=0$, $\mu=0$). (a--e) Probability of ending with everyone using AI (colour scale as in Fig.~\ref{fig:fate}(b); $10$ simulations of $5000$ agents per point), with the boundaries predicted by the first-order (dashed) and second-order (solid) closures. (a,b) $b=0.6$, $\rho=0.4$, with initial adoption $0.6$ and $0.8$; (c) $b=0.75$, $\rho=0.4$, initial adoption $0.42$ (unstable state at $0.36$); (d) $b=0.75$, $\rho=1$, initial adoption $0.57$ (unstable state at $0.53$); (e) $b=0.6$, $\rho=0.4$, initial adoption $0.7$, with initial skills drawn in each group from Beta distributions with means $m_{\rm H}$ and $m_{\rm A}$ and concentration $2$. (f) Boundaries for $b=0.6$, $\lambda_{\rm H}=0.2\lambda_{\rm R}$, initial adoption $0.7$ and three rates of erosion: simulations ($20$ runs per point, logistic fits with standard errors) and second-order closure (lines).}
\label{fig:fateapp}
\end{figure*}

Figure~\ref{fig:fateapp} extends Fig.~\ref{fig:fate}(b) to other initial adoptions and parameters, always for $c=0$ and $\mu=0$. For $b=0.6$ and $\rho=0.4$, with initial adoptions $0.6$ and $0.8$, the second-order closure of Sec.~\ref{sec:fate} locates the boundary within $0.02$ on average, against $0.15$ to $0.16$ for the first-order closure. For $b=0.75$, the initial adoption is chosen slightly above that of the unstable state, $0.36$ for $\rho=0.4$ and $0.53$ for $\rho=1$. The second-order closure then locates the boundary within $0.01$ and $0.05$, respectively, against $0.07$ and $0.18$ for the first-order closure. For $\rho=1$, simulations of $5\times10^4$ agents place the boundary at $p_{\rm H}\simeq0.755$ for $p_{\rm A}=0.6$, against $0.80$ for the second-order closure: the error reflects the closure rather than the size of the population. With Beta-distributed initial skills [panel (e)], the initial variances of the two groups are $m(1-m)/3$, and the second-order closure locates the boundary within $0.025$ on average, against $0.13$ for the first-order closure. With unequal rates [panel (f)], the second-order closure locates the boundaries within $0.02$ on average. For $b=0.6$ with $x_0=0.7$, and for $b=0.75$ with $\rho=1$, replacing the Beta closure for the third moment of non-users by a vanishing third central moment, $\langle s^3\rangle_{\rm H}=m_{\rm H}^3+3m_{\rm H}V_{\rm H}$, changes the predicted boundaries by less than $0.015$.

\begin{figure}[t]
\centering
\includegraphics[width=\columnwidth]{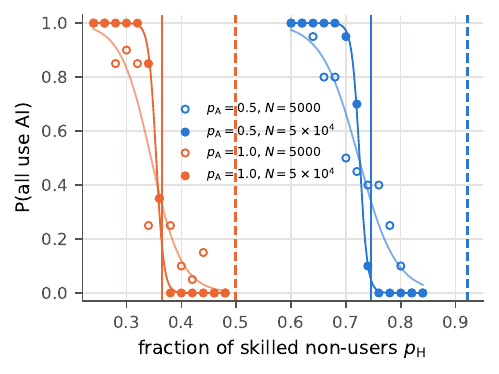}
\caption{\textbf{Size of the population} ($c=0$, $\mu=0$, $b=0.6$, $\rho=0.4$, $x_0=0.7$). Probability of ending with everyone using AI against $p_{\rm H}$, for $p_{\rm A}=0.5$ and $1$, in populations of $5000$ (open symbols) and $5\times10^4$ agents (filled symbols), with $20$ simulations per point and logistic fits (lines), and the boundaries predicted by the first-order (dashed) and second-order (solid) closures (vertical lines).}
\label{fig:size}
\end{figure}

The boundary measured in simulations does not depend on the size of the population (Fig.~\ref{fig:size}). Along the lines $p_{\rm A}=0.5$ and $p_{\rm A}=1$ of Fig.~\ref{fig:fate}(b), when the number of agents increases from $5000$ to $5\times10^4$, the transition between the two outcomes becomes four to five times sharper, but its position moves by less than $0.01$, from $0.723$ to $0.725$ for $p_{\rm A}=0.5$ and from $0.345$ to $0.354$ for $p_{\rm A}=1$, against $0.746$ and $0.365$ for the second-order closure and $0.922$ and $0.499$ for the first-order closure. The errors of the closures are thus not an effect of the size of the population.

\begin{figure}[t]
\centering
\includegraphics[width=\columnwidth]{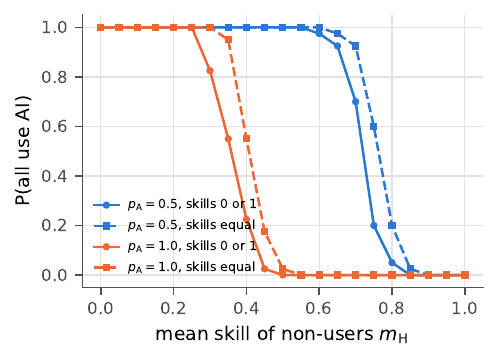}
\caption{\textbf{Initial variance of skills among non-users} ($c=0$, $\mu=0$, $b=0.6$, $\rho=0.4$, $x_0=0.7$). Probability of ending with everyone using AI against the mean skill of non-users, when their skills are initially $0$ or $1$ (solid) or all equal to their mean (dashed), for two values of $p_{\rm A}$ ($40$ simulations of $5000$ agents per point).}
\label{fig:binhomog}
\end{figure}

Figure~\ref{fig:binhomog} compares initial conditions in which non-users have the same mean skill, $m_{\rm H}=p_{\rm H}$, but skills either $0$ or $1$, as in Fig.~\ref{fig:fate}, or all equal to $m_{\rm H}$, for $b=0.6$, $\rho=0.4$ and $x_0=0.7$. The boundaries differ by about $0.04$, binary skills slightly favouring a return, as the unskilled non-users are the first to adopt AI. The second-order closure predicts the boundaries at $m_{\rm H}=0.746$ and $0.785$ for $p_{\rm A}=0.5$, and $0.365$ and $0.421$ for $p_{\rm A}=1$, for binary and homogeneous skills respectively, whereas the first-order closure cannot distinguish them.

\section{Agent-based simulations and robustness}\label{sec:app-abm}

This appendix documents the simulation protocols used to produce the figures and reports robustness checks.

The simulations implement Eqs.~\eqref{eq:returns}--\eqref{eq:imitation} directly for $5000$ to $5\times10^4$ agents. Time advances in steps $\delta t$. In each step, every agent revises with probability $1-e^{-\lambda_{\rm R}\delta t}$, using the payoffs at the start of the step, and skills then relax exactly over the step. This synchronous updating biases adoption by $O(\delta t)$, about $0.01$ at $\delta t=0.1$ and $0.002$ at $\delta t=0.02$, so that quantitative comparisons use $\delta t\le0.05$; the simulations of Figs.~\ref{fig:fate}, \ref{fig:dynamics} and~\ref{fig:fateapp} use $\delta t=0.05$.

\paragraph{Fate of populations.} In Figs.~\ref{fig:fate}, \ref{fig:fateapp}, \ref{fig:size} and~\ref{fig:binhomog} and Table~\ref{tab:predict}, $\mu=0$, and each run continues until all agents, or none, use AI, which every run reached within $1500$ time units. Initial behaviours are assigned to a fraction $x_0$ of agents, and initial skills are drawn independently, equal to $1$ with probability $p_{\rm H}$ for non-users and $p_{\rm A}$ for AI users, and $0$ otherwise. For each value of $p_{\rm A}$, the boundary between the two outcomes is located by fitting $1/\{1+\exp[(p_{\rm H}-p_{\rm H}^*)/\delta]\}$ to the outcomes of the runs by maximum likelihood, which gives its position $p_{\rm H}^*$, a standard error from the curvature of the likelihood, and the width $\delta$ of the transition; rows in which the fraction of runs ending with everyone using AI does not cross $\tfrac12$ are excluded. The curves of Fig.~\ref{fig:size} use $20$ simulations per point, with $p_{\rm H}$ spaced by $0.02$. In Fig.~\ref{fig:fateapp}(e), initial skills are drawn from Beta distributions with parameters $2m_{\rm H}$ and $2(1-m_{\rm H})$ for non-users, and similarly for AI users. In Fig.~\ref{fig:thresholds}(d), $20$ non-users with skill $0$ are placed in a population of $10^4$ AI users with skill $0$, and a run counts as an invasion if adoption falls below $\tfrac12$ before all agents use AI.

\paragraph{Changing benefit.} In Fig.~\ref{fig:dynamics}(b), the population is first equilibrated for $300$ time units at $b=0.1$. The benefit then rises linearly to $1.2$ at a rate $r=3\times10^{-4}$, $10^{-3}$, $3\times10^{-3}$ or $10^{-2}$, i.e.\ over a time $1.1/r$, remains at $1.2$ for $300$ time units, falls back to $0.1$ at the same rate and remains there for $300$ time units. The values of $b$ at which adoption crosses $\tfrac12$ are averaged over three runs. In Fig.~\ref{fig:dynamics}(c), the population is equilibrated for $300$ time units at $b=0.6$, starting from $x=0.01$ with all skills equal to $1$, after which $b=1.2$ for a duration $D$ and then $0.6$ for $400$ time units. We run $20$ simulations for each value of $D$ shown. In Fig.~\ref{fig:dynamics}(d), the same protocol is used with different rates of erosion, and the critical duration $D^*$ is obtained by a logistic fit of the fraction of runs locked in against $D$. A run counts as locked in if its adoption still exceeds $\tfrac12$ after $400$ time units. As $\mu>0$ and $N$ is finite, fluctuations would eventually let a population escape either state, so that `locked in' refers to the duration of the simulations.

\paragraph{Stationary states.} Folds of the stationary curve are located by independent runs at fixed $b$, started in each stable state, with resolution $0.01$ in $b$.

\paragraph{Other imitation rules.} We also tested imitation without decision noise, in which agents copy only peers who earn more, while retaining random exploration. Bistability persists, although the quantitative results differ: for $c=0.15$, $\mu=0.02$ and $\rho=0.39$, the bistable window has width $0.20$, compared with $0.24$ under linear imitation. This rule also produces a persistent core of highly skilled non-users, because those who outperform AI users switch only through exploration. Under the Fermi rule, Eq.~\eqref{eq:zeroflux} gives the stationary density by quadrature. For selection intensity $\gamma=4$, i.e.\ $2.5$ times the selection of the main text ($\gamma=2\beta=1.6$), and $c=0.15$, the bistable window shrinks from $0.41$ at $\rho\approx0.2$ to $0.09$ at $\rho\approx1.2$, as under linear imitation. For the Fermi rule, the folds measured in simulations of $10^4$ agents agree with the stationary solution within their resolution of $0.01$.

\paragraph{Unequal rates of learning and erosion.} When AI users lose skill at rate $\lambda/2$ or $2\lambda$, while non-users gain it at rate $\lambda$, the window for $c=0.15$ and $\lambda=0.2\lambda_{\rm R}$ is $0.23$ in both cases, against $0.24$ for equal rates, in line with Sec.~\ref{sec:rates}. Here too, the folds measured in simulations of $10^4$ agents agree with the stationary solutions within $0.01$.

\paragraph{Tools that complement expertise.} If AI-assisted performance also benefits from the user's skill, $\pi_{\rm A}=b-cx+\eta s$ with $0\le\eta<1$, the feedback towards AI dependence is weakened: the bistable window for $c=0.15$, $\mu=0.02$ and $\rho=0.39$, computed from the exact stationary solution, narrows from $0.24$ at $\eta=0$ to $0.14$ at $\eta=0.2$ and $0.05$ at $\eta=0.4$.

\bibliography{references}

%apsrev4-2.bst 2019-01-14 (MD) hand-edited version of apsrev4-1.bst
%Control: key (0)
%Control: author (8) initials jnrlst
%Control: editor formatted (1) identically to author
%Control: production of article title (0) allowed
%Control: page (0) single
%Control: year (1) truncated
%Control: production of eprint (0) enabled
\begin{thebibliography}{26}%
\makeatletter
\providecommand \@ifxundefined [1]{%
 \@ifx{#1\undefined}
}%
\providecommand \@ifnum [1]{%
 \ifnum #1\expandafter \@firstoftwo
 \else \expandafter \@secondoftwo
 \fi
}%
\providecommand \@ifx [1]{%
 \ifx #1\expandafter \@firstoftwo
 \else \expandafter \@secondoftwo
 \fi
}%
\providecommand \natexlab [1]{#1}%
\providecommand \enquote  [1]{``#1''}%
\providecommand \bibnamefont  [1]{#1}%
\providecommand \bibfnamefont [1]{#1}%
\providecommand \citenamefont [1]{#1}%
\providecommand \href@noop [0]{\@secondoftwo}%
\providecommand \href [0]{\begingroup \@sanitize@url \@href}%
\providecommand \@href[1]{\@@startlink{#1}\@@href}%
\providecommand \@@href[1]{\endgroup#1\@@endlink}%
\providecommand \@sanitize@url [0]{\catcode `\\12\catcode `\$12\catcode
  `\&12\catcode `\#12\catcode `\^12\catcode `\_12\catcode `\%12\relax}%
\providecommand \@@startlink[1]{}%
\providecommand \@@endlink[0]{}%
\providecommand \url  [0]{\begingroup\@sanitize@url \@url }%
\providecommand \@url [1]{\endgroup\@href {#1}{\urlprefix }}%
\providecommand \urlprefix  [0]{URL }%
\providecommand \Eprint [0]{\href }%
\providecommand \doibase [0]{https://doi.org/}%
\providecommand \selectlanguage [0]{\@gobble}%
\providecommand \bibinfo  [0]{\@secondoftwo}%
\providecommand \bibfield  [0]{\@secondoftwo}%
\providecommand \translation [1]{[#1]}%
\providecommand \BibitemOpen [0]{}%
\providecommand \bibitemStop [0]{}%
\providecommand \bibitemNoStop [0]{.\EOS\space}%
\providecommand \EOS [0]{\spacefactor3000\relax}%
\providecommand \BibitemShut  [1]{\csname bibitem#1\endcsname}%
\let\auto@bib@innerbib\@empty
%</preamble>
\bibitem [{\citenamefont {Liu}\ \emph {et~al.}(2026)\citenamefont {Liu},
  \citenamefont {Christian}, \citenamefont {Dumbalska}, \citenamefont
  {Bakker},\ and\ \citenamefont {Dubey}}]{liu}%
  \BibitemOpen
  \bibfield  {author} {\bibinfo {author} {\bibfnamefont {G.}~\bibnamefont
  {Liu}}, \bibinfo {author} {\bibfnamefont {B.}~\bibnamefont {Christian}},
  \bibinfo {author} {\bibfnamefont {T.}~\bibnamefont {Dumbalska}}, \bibinfo
  {author} {\bibfnamefont {M.~A.}\ \bibnamefont {Bakker}},\ and\ \bibinfo
  {author} {\bibfnamefont {R.}~\bibnamefont {Dubey}},\ }\href@noop {} {\bibinfo
  {title} {{AI} assistance reduces persistence and hurts independent
  performance}} (\bibinfo {year} {2026}),\ \Eprint
  {https://arxiv.org/abs/2604.04721} {arXiv:2604.04721 [cs.AI]} \BibitemShut
  {NoStop}%
\bibitem [{\citenamefont {Risko}\ and\ \citenamefont
  {Gilbert}(2016)}]{risko2016}%
  \BibitemOpen
  \bibfield  {author} {\bibinfo {author} {\bibfnamefont {E.~F.}\ \bibnamefont
  {Risko}}\ and\ \bibinfo {author} {\bibfnamefont {S.~J.}\ \bibnamefont
  {Gilbert}},\ }\bibfield  {title} {\bibinfo {title} {Cognitive offloading},\
  }\href {https://doi.org/10.1016/j.tics.2016.07.002} {\bibfield  {journal}
  {\bibinfo  {journal} {Trends Cogn. Sci.}\ }\textbf {\bibinfo {volume} {20}},\
  \bibinfo {pages} {676} (\bibinfo {year} {2016})}\BibitemShut {NoStop}%
\bibitem [{\citenamefont {Bainbridge}(1983)}]{bainbridge1983}%
  \BibitemOpen
  \bibfield  {author} {\bibinfo {author} {\bibfnamefont {L.}~\bibnamefont
  {Bainbridge}},\ }\bibfield  {title} {\bibinfo {title} {Ironies of
  automation},\ }\href {https://doi.org/10.1016/0005-1098(83)90046-8}
  {\bibfield  {journal} {\bibinfo  {journal} {Automatica}\ }\textbf {\bibinfo
  {volume} {19}},\ \bibinfo {pages} {775} (\bibinfo {year} {1983})}\BibitemShut
  {NoStop}%
\bibitem [{\citenamefont {Dahmani}\ and\ \citenamefont
  {Bohbot}(2020)}]{dahmani2020}%
  \BibitemOpen
  \bibfield  {author} {\bibinfo {author} {\bibfnamefont {L.}~\bibnamefont
  {Dahmani}}\ and\ \bibinfo {author} {\bibfnamefont {V.~D.}\ \bibnamefont
  {Bohbot}},\ }\bibfield  {title} {\bibinfo {title} {Habitual use of {GPS}
  negatively impacts spatial memory during self-guided navigation},\ }\href
  {https://doi.org/10.1038/s41598-020-62877-0} {\bibfield  {journal} {\bibinfo
  {journal} {Sci. Rep.}\ }\textbf {\bibinfo {volume} {10}},\ \bibinfo {pages}
  {6310} (\bibinfo {year} {2020})}\BibitemShut {NoStop}%
\bibitem [{\citenamefont {Ying}\ \emph {et~al.}(2024)\citenamefont {Ying},
  \citenamefont {Dong},\ and\ \citenamefont {Fabrikant}}]{ying2024}%
  \BibitemOpen
  \bibfield  {author} {\bibinfo {author} {\bibfnamefont {Q.}~\bibnamefont
  {Ying}}, \bibinfo {author} {\bibfnamefont {W.}~\bibnamefont {Dong}},\ and\
  \bibinfo {author} {\bibfnamefont {S.~I.}\ \bibnamefont {Fabrikant}},\
  }\bibfield  {title} {\bibinfo {title} {How do in-car navigation aids impair
  expert navigators' spatial learning ability?},\ }\href
  {https://doi.org/10.1080/24694452.2024.2356858} {\bibfield  {journal}
  {\bibinfo  {journal} {Ann. Am. Assoc. Geogr.}\ }\textbf {\bibinfo {volume}
  {114}},\ \bibinfo {pages} {1483} (\bibinfo {year} {2024})}\BibitemShut
  {NoStop}%
\bibitem [{\citenamefont {Volz}\ and\ \citenamefont
  {Dorneich}(2020)}]{volz2020}%
  \BibitemOpen
  \bibfield  {author} {\bibinfo {author} {\bibfnamefont {K.~M.}\ \bibnamefont
  {Volz}}\ and\ \bibinfo {author} {\bibfnamefont {M.~C.}\ \bibnamefont
  {Dorneich}},\ }\bibfield  {title} {\bibinfo {title} {Evaluation of cognitive
  skill degradation in flight planning},\ }\href
  {https://doi.org/10.1177/1555343420962897} {\bibfield  {journal} {\bibinfo
  {journal} {J. Cogn. Eng. Decis. Mak.}\ }\textbf {\bibinfo {volume} {14}},\
  \bibinfo {pages} {263} (\bibinfo {year} {2020})}\BibitemShut {NoStop}%
\bibitem [{\citenamefont {Budzy{\'n}}\ \emph {et~al.}(2025)\citenamefont
  {Budzy{\'n}}, \citenamefont {Roma{\'n}czyk}, \citenamefont {Kitala},
  \citenamefont {Ko{\l}odziej}, \citenamefont {Bugajski}, \citenamefont
  {Adami}, \citenamefont {Blom}, \citenamefont {Buszkiewicz}, \citenamefont
  {Halvorsen}, \citenamefont {Hassan}, \citenamefont {Roma{\'n}czyk},
  \citenamefont {Holme}, \citenamefont {Jarus}, \citenamefont {Fielding},
  \citenamefont {Kunar}, \citenamefont {Pellise}, \citenamefont {Pilonis},
  \citenamefont {Kami{\'n}ski}, \citenamefont {Kalager}, \citenamefont
  {Bretthauer},\ and\ \citenamefont {Mori}}]{budzyn2025}%
  \BibitemOpen
  \bibfield  {author} {\bibinfo {author} {\bibfnamefont {K.}~\bibnamefont
  {Budzy{\'n}}}, \bibinfo {author} {\bibfnamefont {M.}~\bibnamefont
  {Roma{\'n}czyk}}, \bibinfo {author} {\bibfnamefont {D.}~\bibnamefont
  {Kitala}}, \bibinfo {author} {\bibfnamefont {P.}~\bibnamefont
  {Ko{\l}odziej}}, \bibinfo {author} {\bibfnamefont {M.}~\bibnamefont
  {Bugajski}}, \bibinfo {author} {\bibfnamefont {H.~O.}\ \bibnamefont {Adami}},
  \bibinfo {author} {\bibfnamefont {J.}~\bibnamefont {Blom}}, \bibinfo {author}
  {\bibfnamefont {M.}~\bibnamefont {Buszkiewicz}}, \bibinfo {author}
  {\bibfnamefont {N.}~\bibnamefont {Halvorsen}}, \bibinfo {author}
  {\bibfnamefont {C.}~\bibnamefont {Hassan}}, \bibinfo {author} {\bibfnamefont
  {T.}~\bibnamefont {Roma{\'n}czyk}}, \bibinfo {author} {\bibfnamefont
  {{\O}.}~\bibnamefont {Holme}}, \bibinfo {author} {\bibfnamefont
  {K.}~\bibnamefont {Jarus}}, \bibinfo {author} {\bibfnamefont
  {S.}~\bibnamefont {Fielding}}, \bibinfo {author} {\bibfnamefont
  {M.}~\bibnamefont {Kunar}}, \bibinfo {author} {\bibfnamefont
  {M.}~\bibnamefont {Pellise}}, \bibinfo {author} {\bibfnamefont
  {N.}~\bibnamefont {Pilonis}}, \bibinfo {author} {\bibfnamefont {M.~F.}\
  \bibnamefont {Kami{\'n}ski}}, \bibinfo {author} {\bibfnamefont
  {M.}~\bibnamefont {Kalager}}, \bibinfo {author} {\bibfnamefont
  {M.}~\bibnamefont {Bretthauer}},\ and\ \bibinfo {author} {\bibfnamefont
  {Y.}~\bibnamefont {Mori}},\ }\bibfield  {title} {\bibinfo {title}
  {Endoscopist deskilling risk after exposure to artificial intelligence in
  colonoscopy: {A} multicentre, observational study},\ }\href
  {https://doi.org/10.1016/S2468-1253(25)00133-5} {\bibfield  {journal}
  {\bibinfo  {journal} {Lancet Gastroenterol. Hepatol.}\ }\textbf {\bibinfo
  {volume} {10}},\ \bibinfo {pages} {896} (\bibinfo {year} {2025})}\BibitemShut
  {NoStop}%
\bibitem [{\citenamefont {Pedreschi}\ \emph {et~al.}(2025)\citenamefont
  {Pedreschi}, \citenamefont {Pappalardo}, \citenamefont {Ferragina},
  \citenamefont {Baeza-Yates}, \citenamefont {Barab{\'a}si}, \citenamefont
  {Dignum}, \citenamefont {Dignum}, \citenamefont {Eliassi-Rad}, \citenamefont
  {Giannotti}, \citenamefont {Kert{\'e}sz}, \citenamefont {Knott},
  \citenamefont {Ioannidis}, \citenamefont {Lukowicz}, \citenamefont
  {Passarella}, \citenamefont {Pentland}, \citenamefont {Shawe-Taylor},\ and\
  \citenamefont {Vespignani}}]{pedreschi2025human}%
  \BibitemOpen
  \bibfield  {author} {\bibinfo {author} {\bibfnamefont {D.}~\bibnamefont
  {Pedreschi}}, \bibinfo {author} {\bibfnamefont {L.}~\bibnamefont
  {Pappalardo}}, \bibinfo {author} {\bibfnamefont {E.}~\bibnamefont
  {Ferragina}}, \bibinfo {author} {\bibfnamefont {R.}~\bibnamefont
  {Baeza-Yates}}, \bibinfo {author} {\bibfnamefont {A.-L.}\ \bibnamefont
  {Barab{\'a}si}}, \bibinfo {author} {\bibfnamefont {F.}~\bibnamefont
  {Dignum}}, \bibinfo {author} {\bibfnamefont {V.}~\bibnamefont {Dignum}},
  \bibinfo {author} {\bibfnamefont {T.}~\bibnamefont {Eliassi-Rad}}, \bibinfo
  {author} {\bibfnamefont {F.}~\bibnamefont {Giannotti}}, \bibinfo {author}
  {\bibfnamefont {J.}~\bibnamefont {Kert{\'e}sz}}, \bibinfo {author}
  {\bibfnamefont {A.}~\bibnamefont {Knott}}, \bibinfo {author} {\bibfnamefont
  {Y.}~\bibnamefont {Ioannidis}}, \bibinfo {author} {\bibfnamefont
  {P.}~\bibnamefont {Lukowicz}}, \bibinfo {author} {\bibfnamefont
  {A.}~\bibnamefont {Passarella}}, \bibinfo {author} {\bibfnamefont {A.~S.}\
  \bibnamefont {Pentland}}, \bibinfo {author} {\bibfnamefont {J.}~\bibnamefont
  {Shawe-Taylor}},\ and\ \bibinfo {author} {\bibfnamefont {A.}~\bibnamefont
  {Vespignani}},\ }\bibfield  {title} {\bibinfo {title} {Human-{AI}
  coevolution},\ }\href {https://doi.org/10.1016/j.artint.2024.104244}
  {\bibfield  {journal} {\bibinfo  {journal} {Artif. Intell.}\ }\textbf
  {\bibinfo {volume} {339}},\ \bibinfo {pages} {104244} (\bibinfo {year}
  {2025})}\BibitemShut {NoStop}%
\bibitem [{\citenamefont {Doshi}\ and\ \citenamefont {Hauser}(2024)}]{doshi}%
  \BibitemOpen
  \bibfield  {author} {\bibinfo {author} {\bibfnamefont {A.~R.}\ \bibnamefont
  {Doshi}}\ and\ \bibinfo {author} {\bibfnamefont {O.~P.}\ \bibnamefont
  {Hauser}},\ }\bibfield  {title} {\bibinfo {title} {Generative {AI} enhances
  individual creativity but reduces the collective diversity of novel
  content},\ }\href {https://doi.org/10.1126/sciadv.adn5290} {\bibfield
  {journal} {\bibinfo  {journal} {Sci. Adv.}\ }\textbf {\bibinfo {volume}
  {10}},\ \bibinfo {pages} {eadn5290} (\bibinfo {year} {2024})}\BibitemShut
  {NoStop}%
\bibitem [{\citenamefont {Meincke}\ \emph {et~al.}(2025)\citenamefont
  {Meincke}, \citenamefont {Nave},\ and\ \citenamefont {Terwiesch}}]{meincke}%
  \BibitemOpen
  \bibfield  {author} {\bibinfo {author} {\bibfnamefont {L.}~\bibnamefont
  {Meincke}}, \bibinfo {author} {\bibfnamefont {G.}~\bibnamefont {Nave}},\ and\
  \bibinfo {author} {\bibfnamefont {C.}~\bibnamefont {Terwiesch}},\ }\bibfield
  {title} {\bibinfo {title} {{ChatGPT} decreases idea diversity in
  brainstorming},\ }\href {https://doi.org/10.1038/s41562-025-02173-x}
  {\bibfield  {journal} {\bibinfo  {journal} {Nat. Hum. Behav.}\ }\textbf
  {\bibinfo {volume} {9}},\ \bibinfo {pages} {1107} (\bibinfo {year}
  {2025})}\BibitemShut {NoStop}%
\bibitem [{\citenamefont {Cornacchia}\ \emph {et~al.}(2026)\citenamefont
  {Cornacchia}, \citenamefont {Nanni}, \citenamefont {Pedreschi},\ and\
  \citenamefont {Pappalardo}}]{cornacchia2026traffic}%
  \BibitemOpen
  \bibfield  {author} {\bibinfo {author} {\bibfnamefont {G.}~\bibnamefont
  {Cornacchia}}, \bibinfo {author} {\bibfnamefont {M.}~\bibnamefont {Nanni}},
  \bibinfo {author} {\bibfnamefont {D.}~\bibnamefont {Pedreschi}},\ and\
  \bibinfo {author} {\bibfnamefont {L.}~\bibnamefont {Pappalardo}},\ }\bibfield
   {title} {\bibinfo {title} {The traffic concentration effects of urban
  navigation services},\ }\href {https://doi.org/10.1038/s41467-026-75254-8}
  {\bibfield  {journal} {\bibinfo  {journal} {Nat. Commun.}\ }\textbf {\bibinfo
  {volume} {17}},\ \bibinfo {pages} {8548} (\bibinfo {year}
  {2026})}\BibitemShut {NoStop}%
\bibitem [{\citenamefont {Karsai}\ \emph {et~al.}(2014)\citenamefont {Karsai},
  \citenamefont {I{\~n}iguez}, \citenamefont {Kaski},\ and\ \citenamefont
  {Kert{\'e}sz}}]{karsai2014complex}%
  \BibitemOpen
  \bibfield  {author} {\bibinfo {author} {\bibfnamefont {M.}~\bibnamefont
  {Karsai}}, \bibinfo {author} {\bibfnamefont {G.}~\bibnamefont {I{\~n}iguez}},
  \bibinfo {author} {\bibfnamefont {K.}~\bibnamefont {Kaski}},\ and\ \bibinfo
  {author} {\bibfnamefont {J.}~\bibnamefont {Kert{\'e}sz}},\ }\bibfield
  {title} {\bibinfo {title} {Complex contagion process in spreading of online
  innovation},\ }\href {https://doi.org/10.1098/rsif.2014.0694} {\bibfield
  {journal} {\bibinfo  {journal} {J. R. Soc. Interface}\ }\textbf {\bibinfo
  {volume} {11}},\ \bibinfo {pages} {20140694} (\bibinfo {year}
  {2014})}\BibitemShut {NoStop}%
\bibitem [{\citenamefont {Szab{\'o}}\ and\ \citenamefont
  {F{\'a}th}(2007)}]{szabo}%
  \BibitemOpen
  \bibfield  {author} {\bibinfo {author} {\bibfnamefont {G.}~\bibnamefont
  {Szab{\'o}}}\ and\ \bibinfo {author} {\bibfnamefont {G.}~\bibnamefont
  {F{\'a}th}},\ }\bibfield  {title} {\bibinfo {title} {Evolutionary games on
  graphs},\ }\href {https://doi.org/10.1016/j.physrep.2007.04.004} {\bibfield
  {journal} {\bibinfo  {journal} {Phys. Rep.}\ }\textbf {\bibinfo {volume}
  {446}},\ \bibinfo {pages} {97} (\bibinfo {year} {2007})}\BibitemShut
  {NoStop}%
\bibitem [{\citenamefont {Castellano}\ \emph {et~al.}(2009)\citenamefont
  {Castellano}, \citenamefont {Fortunato},\ and\ \citenamefont
  {Loreto}}]{castellano}%
  \BibitemOpen
  \bibfield  {author} {\bibinfo {author} {\bibfnamefont {C.}~\bibnamefont
  {Castellano}}, \bibinfo {author} {\bibfnamefont {S.}~\bibnamefont
  {Fortunato}},\ and\ \bibinfo {author} {\bibfnamefont {V.}~\bibnamefont
  {Loreto}},\ }\bibfield  {title} {\bibinfo {title} {Statistical physics of
  social dynamics},\ }\href {https://doi.org/10.1103/RevModPhys.81.591}
  {\bibfield  {journal} {\bibinfo  {journal} {Rev. Mod. Phys.}\ }\textbf
  {\bibinfo {volume} {81}},\ \bibinfo {pages} {591} (\bibinfo {year}
  {2009})}\BibitemShut {NoStop}%
\bibitem [{\citenamefont {Frigerio}\ \emph {et~al.}(2026)\citenamefont
  {Frigerio}, \citenamefont {Caruso}, \citenamefont {Livan},\ and\
  \citenamefont {Latora}}]{frigerio}%
  \BibitemOpen
  \bibfield  {author} {\bibinfo {author} {\bibfnamefont {G.}~\bibnamefont
  {Frigerio}}, \bibinfo {author} {\bibfnamefont {S.}~\bibnamefont {Caruso}},
  \bibinfo {author} {\bibfnamefont {G.}~\bibnamefont {Livan}},\ and\ \bibinfo
  {author} {\bibfnamefont {V.}~\bibnamefont {Latora}},\ }\href@noop {}
  {\bibinfo {title} {Modeling the coevolution of human skills and {AI}
  adoption}} (\bibinfo {year} {2026}),\ \Eprint
  {https://arxiv.org/abs/2610.24794} {arXiv:2610.24794 [physics.soc-ph]}
  \BibitemShut {NoStop}%
\bibitem [{\citenamefont {Park}\ \emph {et~al.}(2026)\citenamefont {Park},
  \citenamefont {Kim},\ and\ \citenamefont {Han}}]{park2026enrichment}%
  \BibitemOpen
  \bibfield  {author} {\bibinfo {author} {\bibfnamefont {J.}~\bibnamefont
  {Park}}, \bibinfo {author} {\bibfnamefont {M.}~\bibnamefont {Kim}},\ and\
  \bibinfo {author} {\bibfnamefont {S.}~\bibnamefont {Han}},\ }\href@noop {}
  {\bibinfo {title} {The enrichment paradox: {C}ritical capability thresholds
  and irreversible dependency in human--{AI} symbiosis}} (\bibinfo {year}
  {2026}),\ \Eprint {https://arxiv.org/abs/2603.24391} {arXiv:2603.24391
  [cs.CY]} \BibitemShut {NoStop}%
\bibitem [{\citenamefont {Sol{\'e}}\ \emph {et~al.}(2026)\citenamefont
  {Sol{\'e}}, \citenamefont {Ruffini}, \citenamefont {Castaldo}, \citenamefont
  {Tuccio}, \citenamefont {Seoane}, \citenamefont {De~Domenico}, \citenamefont
  {Elena}, \citenamefont {Krakauer},\ and\ \citenamefont
  {Levin}}]{sole2026cognitive}%
  \BibitemOpen
  \bibfield  {author} {\bibinfo {author} {\bibfnamefont {R.}~\bibnamefont
  {Sol{\'e}}}, \bibinfo {author} {\bibfnamefont {G.}~\bibnamefont {Ruffini}},
  \bibinfo {author} {\bibfnamefont {F.}~\bibnamefont {Castaldo}}, \bibinfo
  {author} {\bibfnamefont {M.}~\bibnamefont {Tuccio}}, \bibinfo {author}
  {\bibfnamefont {L.~F.}\ \bibnamefont {Seoane}}, \bibinfo {author}
  {\bibfnamefont {M.}~\bibnamefont {De~Domenico}}, \bibinfo {author}
  {\bibfnamefont {S.~F.}\ \bibnamefont {Elena}}, \bibinfo {author}
  {\bibfnamefont {D.~C.}\ \bibnamefont {Krakauer}},\ and\ \bibinfo {author}
  {\bibfnamefont {M.}~\bibnamefont {Levin}},\ }\href@noop {} {\bibinfo {title}
  {Large-language models as a cognitive virus}} (\bibinfo {year} {2026}),\
  \Eprint {https://arxiv.org/abs/2609.03344} {arXiv:2609.03344
  [physics.soc-ph]} \BibitemShut {NoStop}%
\bibitem [{Note1()}]{Note1}%
  \BibitemOpen
  \bibinfo {note} {The present work was developed independently of Ref.~\cite
  {frigerio}, which was posted while this manuscript was being written. Some of
  our results, indicated in the text, were also obtained there in a
  weak-selection approximation, in particular the rate equations, the role of
  the skill of non-users in adoption and the stationary skill of
  non-users.}\BibitemShut {Stop}%
\bibitem [{\citenamefont {Arthur}(1989)}]{arthurlockin}%
  \BibitemOpen
  \bibfield  {author} {\bibinfo {author} {\bibfnamefont {W.~B.}\ \bibnamefont
  {Arthur}},\ }\bibfield  {title} {\bibinfo {title} {Competing technologies,
  increasing returns, and lock-in by historical events},\ }\href
  {https://doi.org/10.2307/2234208} {\bibfield  {journal} {\bibinfo  {journal}
  {Econ. J.}\ }\textbf {\bibinfo {volume} {99}},\ \bibinfo {pages} {116}
  (\bibinfo {year} {1989})}\BibitemShut {NoStop}%
\bibitem [{\citenamefont {Klemperer}(1987)}]{klemperer}%
  \BibitemOpen
  \bibfield  {author} {\bibinfo {author} {\bibfnamefont {P.}~\bibnamefont
  {Klemperer}},\ }\bibfield  {title} {\bibinfo {title} {Markets with consumer
  switching costs},\ }\href {https://doi.org/10.2307/1885068} {\bibfield
  {journal} {\bibinfo  {journal} {Q. J. Econ.}\ }\textbf {\bibinfo {volume}
  {102}},\ \bibinfo {pages} {375} (\bibinfo {year} {1987})}\BibitemShut
  {NoStop}%
\bibitem [{\citenamefont {Traulsen}\ \emph {et~al.}(2005)\citenamefont
  {Traulsen}, \citenamefont {Claussen},\ and\ \citenamefont
  {Hauert}}]{traulsen2005coevolutionary}%
  \BibitemOpen
  \bibfield  {author} {\bibinfo {author} {\bibfnamefont {A.}~\bibnamefont
  {Traulsen}}, \bibinfo {author} {\bibfnamefont {J.~C.}\ \bibnamefont
  {Claussen}},\ and\ \bibinfo {author} {\bibfnamefont {C.}~\bibnamefont
  {Hauert}},\ }\bibfield  {title} {\bibinfo {title} {Coevolutionary dynamics:
  {F}rom finite to infinite populations},\ }\href
  {https://doi.org/10.1103/PhysRevLett.95.238701} {\bibfield  {journal}
  {\bibinfo  {journal} {Phys. Rev. Lett.}\ }\textbf {\bibinfo {volume} {95}},\
  \bibinfo {pages} {238701} (\bibinfo {year} {2005})}\BibitemShut {NoStop}%
\bibitem [{\citenamefont {Traulsen}\ \emph {et~al.}(2007)\citenamefont
  {Traulsen}, \citenamefont {Pacheco},\ and\ \citenamefont {Nowak}}]{traulsen}%
  \BibitemOpen
  \bibfield  {author} {\bibinfo {author} {\bibfnamefont {A.}~\bibnamefont
  {Traulsen}}, \bibinfo {author} {\bibfnamefont {J.~M.}\ \bibnamefont
  {Pacheco}},\ and\ \bibinfo {author} {\bibfnamefont {M.~A.}\ \bibnamefont
  {Nowak}},\ }\bibfield  {title} {\bibinfo {title} {Pairwise comparison and
  selection temperature in evolutionary game dynamics},\ }\href
  {https://doi.org/10.1016/j.jtbi.2007.01.002} {\bibfield  {journal} {\bibinfo
  {journal} {J. Theor. Biol.}\ }\textbf {\bibinfo {volume} {246}},\ \bibinfo
  {pages} {522} (\bibinfo {year} {2007})}\BibitemShut {NoStop}%
\bibitem [{Note2()}]{Note2}%
  \BibitemOpen
  \bibinfo {note} {We have checked numerically that this decrease is monotonic
  for $\rho $ between $0.03$ and $10$.}\BibitemShut {Stop}%
\bibitem [{\citenamefont {Gajos}\ and\ \citenamefont {Mamykina}(2022)}]{gajos}%
  \BibitemOpen
  \bibfield  {author} {\bibinfo {author} {\bibfnamefont {K.~Z.}\ \bibnamefont
  {Gajos}}\ and\ \bibinfo {author} {\bibfnamefont {L.}~\bibnamefont
  {Mamykina}},\ }\bibfield  {title} {\bibinfo {title} {Do people engage
  cognitively with {AI}? {I}mpact of {AI} assistance on incidental learning},\
  }in\ \href {https://doi.org/10.1145/3490099.3511138} {\emph {\bibinfo
  {booktitle} {Proceedings of the 27th International Conference on Intelligent
  User Interfaces (IUI '22)}}}\ (\bibinfo  {publisher} {ACM},\ \bibinfo
  {address} {New York},\ \bibinfo {year} {2022})\ pp.\ \bibinfo {pages}
  {794--806}\BibitemShut {NoStop}%
\bibitem [{\citenamefont {Bastani}\ \emph {et~al.}(2025)\citenamefont
  {Bastani}, \citenamefont {Bastani}, \citenamefont {Sungu}, \citenamefont
  {Ge}, \citenamefont {Kabakc{\i}},\ and\ \citenamefont {Mariman}}]{bastani}%
  \BibitemOpen
  \bibfield  {author} {\bibinfo {author} {\bibfnamefont {H.}~\bibnamefont
  {Bastani}}, \bibinfo {author} {\bibfnamefont {O.}~\bibnamefont {Bastani}},
  \bibinfo {author} {\bibfnamefont {A.}~\bibnamefont {Sungu}}, \bibinfo
  {author} {\bibfnamefont {H.}~\bibnamefont {Ge}}, \bibinfo {author}
  {\bibfnamefont {{\"O}.}~\bibnamefont {Kabakc{\i}}},\ and\ \bibinfo {author}
  {\bibfnamefont {R.}~\bibnamefont {Mariman}},\ }\bibfield  {title} {\bibinfo
  {title} {Generative {AI} without guardrails can harm learning: {E}vidence
  from high school mathematics},\ }\href
  {https://doi.org/10.1073/pnas.2422633122} {\bibfield  {journal} {\bibinfo
  {journal} {Proc. Natl. Acad. Sci. U.S.A.}\ }\textbf {\bibinfo {volume}
  {122}},\ \bibinfo {pages} {e2422633122} (\bibinfo {year} {2025})}\BibitemShut
  {NoStop}%
\bibitem [{\citenamefont {Sparrow}\ \emph {et~al.}(2011)\citenamefont
  {Sparrow}, \citenamefont {Liu},\ and\ \citenamefont {Wegner}}]{sparrow2011}%
  \BibitemOpen
  \bibfield  {author} {\bibinfo {author} {\bibfnamefont {B.}~\bibnamefont
  {Sparrow}}, \bibinfo {author} {\bibfnamefont {J.}~\bibnamefont {Liu}},\ and\
  \bibinfo {author} {\bibfnamefont {D.~M.}\ \bibnamefont {Wegner}},\ }\bibfield
   {title} {\bibinfo {title} {Google effects on memory: {C}ognitive
  consequences of having information at our fingertips},\ }\href
  {https://doi.org/10.1126/science.1207745} {\bibfield  {journal} {\bibinfo
  {journal} {Science}\ }\textbf {\bibinfo {volume} {333}},\ \bibinfo {pages}
  {776} (\bibinfo {year} {2011})}\BibitemShut {NoStop}%
\end{thebibliography}%

\end{document}